\documentclass[manuscript,screen]{acmart}

\AtBeginDocument{%
  \providecommand\BibTeX{{%
    \normalfont B\kern-0.5em{\scshape i\kern-0.25em b}\kern-0.8em\TeX}}}

\setcopyright{acmcopyright}
\copyrightyear{2018}
\acmYear{2022}
\acmDOI{XXXXXXX.XXXXXXX}

\acmPrice{15.00}
\acmISBN{978-1-4503-XXXX-X/18/06}

\usepackage{color}
\usepackage{xcolor}
\usepackage{booktabs}
\usepackage{textcomp}

\usepackage{pifont}% http://ctan.org/pkg/pifont
\newcommand{\cmark}{\ding{51}}%
\newcommand{\xmark}{\ding{53}}%

\usepackage{lipsum}
\usepackage{caption}
\usepackage[utf8]{inputenc}
\usepackage{pgfplots}
\DeclareUnicodeCharacter{2212}{−}
\usepgfplotslibrary{groupplots,dateplot}
\usetikzlibrary{patterns,shapes.arrows}
\pgfplotsset{compat=newest}
\usepackage[mode=buildnew]{standalone}% requires -shell-escape
\usepackage{tikz}
\usetikzlibrary{fit,shapes,backgrounds,arrows}
\usetikzlibrary{shapes.multipart}

\usepackage[english]{babel}
\usepackage[autostyle,english=american]{csquotes}

\begin{document}

\title{Predictability and Comprehensibility in Post-Hoc XAI Methods: A User-Centered Analysis}

\author{Anahid Jalali}
\email{anahid.jalali@ait.ac.at}
\affiliation{%
  \institution{AIT - Austrian Institute of Technology}
  \city{Vienna}  
  \country{Austria}
}

\author{Bernhard Haslhofer}
\affiliation{%
  \institution{AIT - Austrian Institute of Technology}
  \city{Vienna}
  \country{Austria}
}
\email{bernhard.haslhofer@ait.ac.at}

\author{Simone Kriglstein}
\affiliation{%
  \institution{AIT - Austrian Institute of Technology}
  \city{Vienna}
  \country{Austria}
}
\affiliation{
  \institution{Masaryk University}
  \city{Brno}
  \country{Czech Republic}
}
\email{simone.kriglstein@ait.ac.at}

\author{Andreas Rauber}
\affiliation{%
  \institution{Vienna University of Technology}
  \city{Vienna}
  \country{Austria}
}
\email{andreas.rauber@ifs.tuwien.ac.at}


\renewcommand{\shortauthors}{Jalali, et al.}

\begin{abstract}
    % What is the problem / motivation for our work (max 1-2 sentences)
    Post-hoc explainability methods aim to clarify predictions of black-box machine learning models. However, it is still largely unclear how well users comprehend the provided explanations and whether these increase the users' ability to predict the model behavior. 
    % What is our contribution to solving that problem / our research questions?  (max 1-2 sentences)
    We approach this question by conducting a user study to evaluate comprehensibility and predictability in two widely used tools: LIME and SHAP. Moreover, we investigate the effect of counterfactual explanations and misclassifications on users' ability to understand and predict the model behavior.
    % What did we find?  (max 1-2 sentences)
    We find that the comprehensibility of SHAP is significantly reduced when explanations are provided for samples near a model's decision boundary. Furthermore, we find that counterfactual explanations and misclassifications can significantly increase the users' understanding of how a machine learning model is making decisions.
    % What is the impact of our findings  (max 1 sentence)
    Based on our findings, we also derive design recommendations for future post-hoc explainability methods with increased comprehensibility and predictability.
\end{abstract}

%%%%%%%%%%%%%%%% User Study HCI? %%%%%%%%%
\begin{CCSXML}
<ccs2012>
<concept>
<concept_id>10003120.10003121.10003122.10003334</concept_id>
<concept_desc>Human-centered computing~User studies</concept_desc>
<concept_significance>500</concept_significance>
</concept>
</ccs2012>
\end{CCSXML}

\ccsdesc[500]{Human-centered computing~User studies}

%%%%%%%%%%%%%%%% Or ML ?? %%%%%%%%%
%\begin{CCSXML}
%<ccs2012>
%<concept>
%<concept_id>10010147.10010257</concept_id>
%<concept_desc>Computing methodologies~Machine learning</concept_desc>
%<concept_significance>500</concept_significance>
%</concept>
%</ccs2012>
%\end{CCSXML}

%\ccsdesc[500]{Computing methodologies~Machine learning}

\keywords{Interpretable Machine Learning, XAI, XAI Evaluation, User Study}

\settopmatter{printfolios=true}

\maketitle

% !TeX root = ../AAAI-2022.tex

\section{Introduction}

% Motivation & Problem statement
The opacity of machine learning models is a well-known problem in application areas such as health care systems, financial services, or industrial applications~\cite{carvalho2019machine}, where transparency and accountability are fundamental requirements.
When models are not transparent, users and machine learning experts (ML) have difficulty explaining how models arrive at their predictions. Therefore, ongoing research in the field of Interpretable Machine Learning (IML), also known as eXplainable Artificial Intelligence (XAI), focuses on implementing methods that either examine the inner structure of models (model-specific) or explain predictions of a trained model based on a training dataset (post-hoc)~\cite{molnar2020interpretable}.

Well-known post-hoc explanation techniques are: Local Interpretable Model Agnostic Explanations (LIME)~\cite{ribeiro2016should}, DeepLift~\cite{shrikumar2017learning}, COVAR~\cite{haufe2014interpretation},
%which is a simple univariate measure of covariance between features and model predictions,
and Shapely Additive exPlanations (SHAP)~\cite{lundberg2017unified}. They provide different types of explanations to the user and aim at fulfilling the goals of interpretability, which are: following a model's prediction for a given dataset, being easy to comprehend, and being efficient~\cite{ruping2006learning}. 
However, as~\cite{carvalho2019machine} stated, there is ambiguity in the definition of interpretability and how it should be measured and evaluated. The authors further argued that application-grounded evaluation is the most appropriate as \textquote{it assesses interpretability in the end goal with the end-users}. 

% Related work
Several user studies already evaluated model interpretations and explanations: ~\cite{ribeiro2016should,ribeiro2018anchors} measured comprehensibility and trust by asking users to explain the best model, the most suitable features, as well as model behaviors and irregularities. Other studies also focused on measuring  comprehensiveness~\cite{nourani2019effects,papenmeier2019model}, usefulness~\cite{ribeiro2016should,mohseni2018human},
and trustworthiness of explanations~\cite{papenmeier2019model,nourani2019effects,schmidt2019quantifying,lakkaraju2020fool}. However, they all evaluated a single XAI method and aimed to improve its accuracy for a particular task or manipulated explanations to mislead users and measure their bias when presenting a fidelity or model accuracy score for a given task. Furthermore, they studied mainly annotation tasks on image or text datasets.

% Why is the work of Schmidt et al. insufficient
Our work is motivated by~\cite{schmidt2019quantifying}, who measured comprehensibility and trust based on the users' interaction time in a text classification task. However, it remains unclear how users make their judgments: do they blindly follow provided model explanations or make their judgments based on the real-world meaning of data points, which were words in this case? Furthermore, current studies do not infer recommendations that could inform future, improved model explainability methods.

The need for XAI user studies has been pointed out by ~\cite{kaur2020interpreting}, who argue that XAI techniques must align with the mental model of ML-practitioners. Also, ~\cite{mohseni2019toward} stated that \textquote{the ultimate goal is for people (experts and/or users) to understand the models, and it is, therefore, essential to involve human feedback and reasoning as a requisite component for design and evaluation of interpretable-ML systems.}

% Our Aims & Contributions
We acknowledge the work of Jacovi et al.~\cite{jacovi2021formalizing} on formalizing human-AI trust, in which they stated \textit{"defining trust as the user's attempt to predict the impact of the model behavior under risk and uncertainty~\cite{hoffman2017taxonomy} is a goal but not necessarily a symptom of trust"}. Moreover, Mohseni et al.~\cite{mohseni2021multidisciplinary}, defines one of the desired properties of explainer systems as ``predictability'', which is the ability of these systems to \textquote{support building a mental model of the system that enables user to predict system behavior.}
Therefore, in this work, we aim at measuring the \emph{comprehensibility} and the \emph{predictability}, to compare two well-known and widely used XAI approaches, SHAP and LIME. For this purpose, we first refine our notion of \emph{comprehensibility} and \emph{predictability} in classification tasks as follows:

\textbf{Comprehensibility}: denotes the user's ability to transfer information on feature contributions obtained from model explanations across samples of the same class.

Previous studies~\cite{guidotti2018survey, ribeiro2016should, schmidt2019quantifying, lipton2018mythos} assume trust as \textquote{trust in model correctness} and evaluate the user's ability to guess a sample's label correctly, given model explanations. We additionally consider both notions of predictability~\cite{mohseni2021multidisciplinary} and simulation~\cite{hase2020evaluating}, which is the user's ability to guess a model's prediction on a new sample correctly, and broaden the definition of predictability as follows.

\textbf{Predictability}: denotes the explainer's ability to support the users with predicting model predictions, which can be correct or incorrect, on a new sample given model explanations for a sample that the model predicted correctly with high confidence.

In the following, we aim to evaluate SHAP and LIME explanations qualitatively and quantitatively as part of a user study. We formulate our research questions as follows:

\begin{displayquote}
    \textbf{RQ1}. To what extent do users comprehend the explanations provided by different XAI methods, and are they able to predict the decision made by the model?
\end{displayquote}

When interacting with these tools, we noticed that it is essential to understand how features impact a model decision and that this is easier to comprehend when a model is more confident about a decision. We tested these hypotheses and answered the above question in a comparative study, which we describe in more detail in Section \emph{RQ1: Comprehensibility}. We found evidence that supports our hypotheses for SHAP.

Furthermore, we observed that users who understand model predictions for the two different classes found it easier to classify unexplained and unlabeled samples. Therefore, we hypothesize that users predictability increases when they can classify new samples using the explanations from samples of different classes. In Section \emph{RQ1: Predictability}, we elaborate on this in more detail and show that users are able to predict the decisions made by the model, using both SHAP and LIME equally.

Given these findings, we further investigate the effect of counterfactual and misclassified samples on the users ability to predict the model's decision. More precisely, we consider the following research question:

\begin{displayquote}
    \textbf{RQ2}. To what extent can visualizations of counterfactual and misclassified samples improve the user's predictability? 
\end{displayquote}

We answered this question by testing the following hypothesis: adding explanations of misclassified and counterfactual samples can improve the predictability and support anticipating the model's behavior. We describe our experiment in more details in Section \emph{RQ2: Improving Predictability with Visualizations} and report evidence that supports that hypothesis for both SHAP and LIME. We found that users have higher predictability with LIME than with SHAP explanations.

Moreover, throughout our experiments, we asked users to provide their subjective feedback on given explanations. Their responses allowed us to answer our third and final research question: 

\begin{displayquote}
    \textbf{RQ3}. To what extent can visualizations of local XAI explanations guide users in finding global explanations?
\end{displayquote}

Throughout our experiments, we collected user feedback for both LIME and SHAP, and asked the participants to explain possible shortcomings and sources of confusion. As detailed in Section \emph{RQ3: Qualitative Analysis}, we observed that negative feedback provides valuable insight into how users interpret explanations. For example, the users complained about inconsistent explanations of values and different scaling of visualizations. Our experiments showed that this reduced the ability of the user to understand the behavior of a model.

The findings of our comparative user study can substantially contribute to the design of new or refinement of existing XAI methods, and therefore, we propose a set of design recommendations in Section \emph{Discussion}, which we will confer in more detail.
% !TeX root = ../AAAI-2022.tex

\section{Related Work}
\label{sota}

% Evaluation criteria and baselines
Previous studies, such as ~\cite{carvalho2019machine}, ~\cite{zhou2021evaluating}, ~\cite{tonekaboni2019clinicians} or ~\cite{sokol2020explainability}, discuss the various properties of explainability and define evaluation criteria for qualitative (e.g. measuring explanations' comprehensiveness, trustworthiness) and quantitative (e.g. measuring explanations' accuracy, fidelity and consistency) approaches.

The need for an explainability baseline for evaluating the quality of explanations was raised by ~\cite{mohseni2018human}, who measured the trustworthiness of explanations and quantitatively compared LIME with Grad-Cam explanations on three different baselines: human-attention mask, segmentation mask, and human judgment ratings. Their findings indicate human biases in ratings and significant differences in evaluation scores, which they assume to be caused by ``clear non-uniform distribution of weights in human attention masks''.

Our work builds on previous user studies, which evaluated model interpretations and explanations. We summarize them into Table~\ref{table:sota} and roughly divide them into three categories: the first category (\cite{alqaraawi2020evaluating,papenmeier2019model,schmidt2019quantifying}) evaluates explainability quantitatively and measures the success of the user with and without explanations. The second category (\cite{hase2020evaluating,nourani2019effects,papenmeier2019model, lakkaraju2020fool,ribeiro2016should,ribeiro2018anchors} measures comprehensiveness and trustworthiness of explanations quantitatively by aligning meaningful and meaningless or manipulated explanations with human logic. They point out that manipulated explanations can increase user trust in biased models and lead to mistrust when explanations do not match the decision-making process of users. The third category (\cite{kaur2020interpreting,shin2021effects,spinner2019explainer, wang2019designing}) measures the comprehensibility and trust of users by considering system interactions. These approaches focus on understanding how users perceive information, diagnose and refine the AI systems. Their qualitative results are often used as recommendations for designing the XAI approaches to improve users' trust. 

Thus, previous studies focused mainly on image or text data, misleading the users with manipulated explanations in binary annotation tasks. User considerations of the decision-making process and design recommendations on improving explanations were mainly out of scope. For more detailed information on each of the mentioned works, please refer to the summary in table~\ref{table:sota}.

% Our delta and contribution
Therefore, we do not limit ourselves to a quantitative evaluation but also evaluate XAI approaches qualitatively and consider user feedback to understand the reasons behind possible confusion in decision making. Moreover, we include explanations of counterfactual and misclassified samples to test users' predictability using the explanations.

\begin{table}[h]
    \centering
    \caption{A summary of existing studies for evaluating XAI approaches.}
    \label{table:sota}
     \scalebox{0.80}{ % remove the scaling for a double column template
    \begin{tabular}{cccccccc}
        \toprule
        &&\multicolumn{2}{c}{Evaluation}&\multicolumn{4}{c}{Explanations}\\
        \cmidrule{2-8}
        Paper&Metric&Approach&Data&\multicolumn{2}{c}{Manipulated}&\multicolumn{2}{c}{Show}\\
        \cmidrule{5-8}
        &&&&Model&Examples&Preds&Labels\\
        \midrule

        %Alqaraawi et al.
        ~\cite{alqaraawi2020evaluating}&Compreh.&Quan.&Image&\xmark&\xmark&\cmark&\xmark\\
        \midrule

        %Hase et al.
        ~\cite{hase2020evaluating}&Simulation&Quan.&Text&\xmark&\cmark&\cmark&\xmark\\
                                             &(Trust)&&&&&\\
        % This work is also measures model behaviour prediction using explanations, but they don't call it trust ... they call it simulation ...
        \midrule

        %Kaur et al.
        ~\cite{kaur2020interpreting}&Trust&Quan.&Tabular&\xmark&\xmark&\cmark&\xmark\\
                                               &&Qual.&&&&\\
        \midrule
        
        %Lakkaraju et al.
        ~\cite{lakkaraju2020fool}&Trust&Quan.&Tabular&\cmark&\cmark&\xmark&\xmark\\
        \midrule
        
        %Mohseni et al.
        ~\cite{mohseni2018human}&Trust&Quan.&Image&\xmark&\xmark&\cmark&\xmark\\
                                 &&&Text&&&\\
        \midrule
        
        %Nourani et al.
        ~\cite{nourani2019effects}&Trust&Quan.&Image&\xmark&\cmark&\xmark&\xmark\\
        \midrule
        
        %Papenmeier et al.
        ~\cite{papenmeier2019model}&Trust&Quan.&Text&\xmark&\cmark&\xmark&\xmark\\
                                   &Compreh.&&&&&\\
        \midrule
        %Ribeiro et al.
        ~\cite{ribeiro2016should}&Trust&Quan.&Image&\cmark&\xmark&\cmark&\xmark\\
                                 &Compreh.&&Text&&&\\
        \midrule
        %Ribeiro et al.
        ~\cite{ribeiro2018anchors}&Compreh.&Quan.&Image&\xmark&\xmark&\cmark&\xmark\\
        
        \midrule
        %Schmidt et al.
        ~\cite{schmidt2019quantifying}&Trust&Quan.&Text&\cmark&\xmark&\cmark&\xmark\\
                                      &Compreh.&&&&&\\
        \midrule                             
        
        %Shin
        ~\cite{shin2021effects}&Trust&Quan.&Text&\xmark&\xmark&\cmark&\xmark\\
        &&&&&&\\
        \midrule
        
        %Spinner et al.
        ~\cite{spinner2019explainer}&Compreh.&Qual.&Image&\xmark&\xmark&\cmark&\xmark\\
        \midrule
        
        %Wang et al.
        ~\cite{wang2019designing}&Trust&Qual.&Tabular&\xmark&\xmark&\cmark&\xmark\\
        &&&(Medical)&&&\\
        \midrule
        
        Our Work&Predictability&Quan.&Tabular&\xmark&\xmark&\cmark&\cmark\\
                &Compreh.&Qual.&&&&\\        
        
        \bottomrule 
    \end{tabular}
    }
\end{table}
% !TeX root = ../AAAI-2022.tex

\section{Methodology}

We conducted a user study to evaluate comprehensibility and predictability in explanations provided by two widely-used XAI methods: SHAP and LIME. Our experimental setup follows a between-subject design, with the XAI method as the primary varying condition. That means we exposed each participant to only one condition (LIME or SHAP), which shortened the duration of experimental sessions for each participant (c.f.,~\cite{martin2007}).

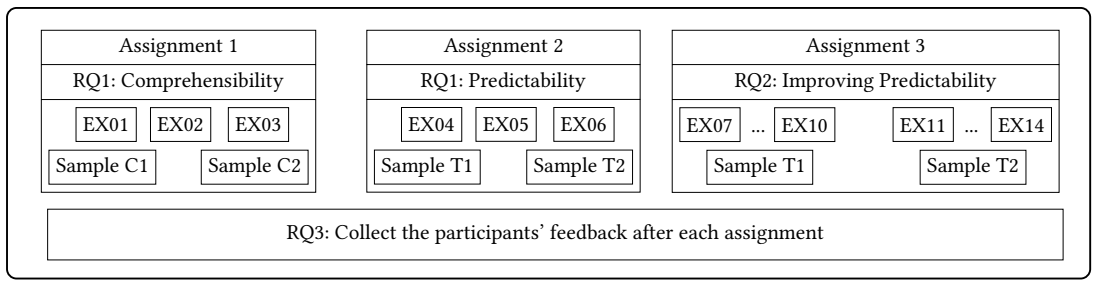
\begin{figure*}
  \centering % remove scaling for a double column template
  \scalebox{0.9}{\begin{tikzpicture}[
  triple/.style={draw, anchor=text, rectangle split,rectangle split parts=3}]
  
  \tikzstyle{bigbox} = [minimum width=16cm, minimum height=4cm,draw, thick, rounded corners, rectangle]
  \tikzstyle{box} = [minimum width=0.5cm, minimum height=0.5cm,rectangle, draw]
  \tikzstyle{whitebox} = [minimum width=0.5cm, minimum height=0.5cm,rectangle]
  \tikzstyle{longbox} = [minimum width=15cm, minimum height=0.75cm,rectangle, draw]
  \tikzset{txtstyle/.style={text=black}}

\node[triple][xshift=-7.5cm, yshift=0.95cm] {Assignment 1
    \nodepart{second}
      {RQ1: Comprehensibility}
    \nodepart{third}
      \tikz{
      \node[box, xshift=-1.35cm, yshift=3.5cm] {EX01};
      \node[box, xshift= -0.25cm, yshift=3.5cm] {EX02};
      \node[box, xshift= 0.90cm, yshift=3.5cm] {EX03};
      
      \node[box, xshift=-1.75cm, yshift=2.9cm] {Sample C1};
      \node[box, xshift=0.5cm, yshift=2.9cm] {Sample C2};
      }
  };
  
\node[triple][xshift=-2.7cm, yshift=0.95cm] {Assignment 2
    \nodepart{second}
      {RQ1: Predictability}
    \nodepart{third}
      \tikz{
      \node[box, xshift=-1.35cm, yshift=3.5cm] {EX04};
      \node[box, xshift= -0.25cm, yshift=3.5cm] {EX05};
      \node[box, xshift= 0.90cm, yshift=3.5cm] {EX06};
      
      \node[box, xshift=-1.75cm, yshift=2.9cm] {Sample T1};
      \node[box, xshift=0.5cm, yshift=2.9cm] {Sample T2};
      }
  };
  
\node[triple][xshift=2.65cm, yshift=0.95cm] {Assignment 3
    \nodepart{second}
      {RQ2: Improving Predictability}
    \nodepart{third}
      \tikz{
      \node[box, xshift=-1.15cm, yshift=3.5cm] {EX07};
      \node[whitebox, xshift= -0.2cm, yshift=3.5cm] {...};
      \node[box, xshift= 0.25cm, yshift=3.5cm] {EX10};
      
      \node[box, xshift=2cm, yshift=3.5cm] {EX11};
      \node[whitebox, xshift= 2.95cm, yshift=3.5cm] {...};
      \node[box, xshift= 3.45cm, yshift=3.5cm] {EX14};
      
      \node[box, xshift=-0.75cm, yshift=2.9cm] {Sample T1};
      \node[box, xshift=2.4cm, yshift=2.9cm] {Sample T2};
      }
  };
  
 % Qualitative data
    \node[longbox][xshift=-1.05cm, yshift=-1.75cm]
    {RQ3: Collect the participants' feedback after each assignment};

% Background box
  \begin{pgfonlayer}{background}
    \node[bigbox][xshift=-1.15cm, yshift=-0.4cm] {};
  \end{pgfonlayer} 

  \end{tikzpicture}}
  \caption[An overview of the assignments of our user study]{An overview of the assignments of our user study. We asked participants to work on assignments and specific tasks, answering a specific research question in the main survey. Assignment one and two contained three samples and their explanations, depicted as \emph{EXxx}, provided by either LIME or SHAP. The users studied the samples and explanations and had to answer questions for a test sample, depicted as \emph{Sample Cx} or \emph{Tx}}.
  \label{fig:survey_flow}
\end{figure*}

Figure~\ref{fig:survey_flow} depicts our overall experimental setup, which started with participant recruitment and a pre-test survey phase. As part of the central survey, we defined, for each research question, several assignments and tasks to be solved by the participants. The first assignment measured the user's comprehensibility for a given XAI method, and the second one the explainer's predictability. The third assignment investigated the effect of adding explanations of misclassified and counterfactual samples on explainer's predictability. 
In each task, we presented visualizations of SHAP- or LIME-explanations to the participant and asked them to answer four questions, in which they had to interpret the visualizations for the given test sample. %An example of LIME and SHAP's visualizations are illustrated in Figure~\ref{fig:RQ1_LIME} and Figure~\ref{fig:RQ1_trust_SHAP}, respectively.

\paragraph{Dataset and Implementation.}

We chose the Boston Housing dataset~\cite{harrison1978hedonic} because of its simplicity and transparency. This dataset estimates the median price of apartments in Boston. We transferred this regression task into a classification task by categorizing the estimated prices into three classes: 1) low-price, 2) medium-price, and 3) high-price while preserving the feature correlations with the target variable. We used five features that give our model the highest accuracy: average number of rooms, pupil and teacher ratio, air pollution level, crime rate, and the zone where an apartment is located.

In an initial trial experiment, we found that participants tend to project their interpretation of feature labels (e.g., crime rate) onto an explanation instead of interpreting the information provided by either LIME or SHAP. Therefore, we anonymized the feature names to \textit{F1, F2, F3, F4}, and \textit{F5}, respectively.

We further min-max normalized each feature and trained a machine learning model using a 3-layered fully connected dense neural network (64 units, Relu function, and a softmax at the output layer). We optimized the model's trained weights with Stochastic Gradient Descent (0.001 learning rate). The model had 93\% accuracy, and median prediction probability of 70.21\%, 46.81\%, and 72.87\% for low-price, medium-price, and high-price classes, respectively. However, we did not provide the participants with this information.

To set up our experiment, we used Python 3.7, and for the explainability visualizations from LIME and SHAP, we used the \textit{LimeTabularExplainer} (\textit{lime} library version 0.2.0.1), and the \textit{KernelExplainer} (\textit{shap} library version 0.34.0). 

\paragraph{Participants.}

We recruited participants with ML and Data Science experience having technical backgrounds in Computer Science, Mathematics, and Physics. The participants were scientists and practitioners from eight different institutions in five countries, collected via their LinkedIn profile. Overall, 47 participants took part in our experiment, and we randomly assigned one XAI approach, either SHAP or LIME, to each participant. We compensated participants with €20 for their approximately 1-hour effort of taking part in the experiment. 

We had to remove one user, who answered ``I do not know'' and stated afterwards that he was not focused and could not participate in this study. This data cleaning step left us with a total of 46 participants: 30 male and 16 female with an average age of 31, 24 users evaluated LIME, and 22 users evaluated SHAP. When asked about their experience with explainability approaches and interpreting machine learning; they often stated that they interpret models by looking at feature importance plots of decision trees or on the coefficients of linear regression models. On the other hand, only twelve users had experience with XAI approaches such as LIME, Layer-wise relevance-propagation (LRP), heatmaps, or Google's Language Interpretability Tool (LIT)~\cite{tenney2020language}. Therefore, we can assume that most of our participants (34 of 46) were non-experts in XAI.

\paragraph{Survey Procedure.}

We implemented our survey using the~\cite{Qualtrics} platform. After the users gave their consent to the overall experimental design, they started the pre-test survey, in which we asked them about their demographic information, background, and data science experience. Then, we measured their experience by presenting them with 12 data science and machine learning know-how questions such as bias and variance trade-off or distribution functions.

Afterwards, we acquainted the participants with the overall survey procedure in an initial training phase, in which we explained the dataset, the tasks, the visualizations provided by each explainability approach, and the structure of the assignments. Then, in the second part of the survey, the participants started working on the three assignments, each comprising two tasks with four questions. Thus, each participant had to answer 24 questions related to interpretability in total. 

Finally, we present the participants with the Nasa Task Load Index (TLX)~\cite{hart1988development} questionnaire to obtain insight into the mental, physical and temporal demand of the survey as well as the participants' success, effort, and frustration. 

\paragraph{Data Coding.}

For the quantitative part of our analysis, we assigned scores to each multiple-choice answer and computed the sum of all answers to compare assignment results. We gave each correct answer a score of 2, each wrong answer a -1, and "I do not know" answers a 0. This scoring scheme allows us to distinguish participants who tried to answer the questions seriously but possibly wrong from those who just checked ``I don't know.''

For the qualitative evaluation, we interviewed the participants and transcribed their feedback throughout each interview session. We followed the Mayring qualitative analysis decoding rules described in~\cite{mayring2004qualitative} to categorize the transcribed participant feedback. %We describe the details of our coding rules in Appendix~\ref{app:coding_rules}.
% !TeX root = ../AAAI-2022.tex

\section{RQ1: Comprehensibility}
\label{RQ1_H1}

% Goal
This section provides answers to our first research question, RQ1, which seeks to understand the relationship between the comprehensibility of a set of three explanations for the user and the prediction confidence of a machine learning model. 

\begin{figure}[t]
  \centering % double column scale to 0.97
  \scalebox{0.65}{\input{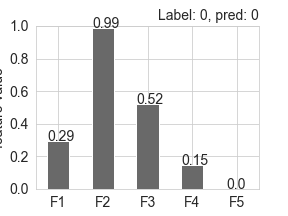}}
  \caption{Example explanation (EX01) provided to the user. On the left hand side, it shows a sample (an apartment) that the model classified as being low-price (label 0). The x-axis of the bar plot are the sample's attributes and the y-axis are the values. On the right hand side, a LIME explanation, describes the decision of the model.}
  \label{fig:RQ1_LIME}
\end{figure}

% Method
For this purpose, we randomly assigned each user to a XAI method (LIME or SHAP) and presented them with three explained samples (EX01, EX02, and EX03), each representing an apartment that the machine learning model classified as belonging to the class \emph{low-price} (coded as 0). Figure~\ref{fig:RQ1_LIME} illustrates one of these three samples and shows how we presented and explained it to the users. We have chosen these samples based on their information and ensured that several features contributed to its low-price classification.

Moreover, we chose two additional test samples from the Boston Housing dataset, C1 and C2, with features similar to the explained samples, which the model correctly classified as low-price. However, the prediction confidence for C1 was higher than the confidence for C2, indicating that C1 is further away from the model's decision boundary. We use the model's Probability Distribution Delta (PDD) to quantify the model's confidence.

The users studied the explanations and tried to use the information they learned from EX01, EX02, and EX03 to answer the following multiple-choice questions for C1 and then C2:

\begin{enumerate}
    \item Choose two features that highly influence the prediction of class ``low-price (0)''.
    \item How does the value of F2 (together with F1 and F5) influence the model's decision on class ``low-price (0)''?
    \item How does the value of F1 affect the probability of class ``low-price (0)''?
    \item How does the value of F3 in this sample (w.r.t. explanations) affect the probability of class ``low-price (0)''?
\end{enumerate}

With the above questions, we wanted to measure the user's understanding of whether they can interpret how feature values can increase or decrease the probability of being classified as ``low price''. Therefore, we also formulated control questions to avoid random answers and ensure participants focused on the tasks. For the third question, for instance, the answer should match the answer to the second question. If this is not the case, we know that an answer is random and that we should remove it from our analysis. However, this was not the case with our users, and they did not randomly answer the questions.

We compute scores for all the responses, giving an overall comprehensibility score for each user and each assignment. We also compared the individual question scores of C1 and C2 to measure quantitatively whether the presented visualization helps the user comprehend each feature's contribution to the model's decision.

Furthermore, we code the participant's interview responses into three categories: (i) C1 was more difficult than C2, (ii) C2 was more difficult than C1, and (iii) C1 and C2 were equally challenging. %We provide further details on our coding procedure in Appendix~\ref{Appx_comprehensibility}.

\paragraph{Results}

\begin{figure}[t]
    \centering 
    % This file was created by tikzplotlib v0.9.7.
\begin{tikzpicture}[scale=0.8]

\definecolor{color0}{rgb}{0.795098039215686,0.200980392156863,0.206862745098039}
\definecolor{color1}{rgb}{0.278921568627451,0.487745098039216,0.658333333333333}
\definecolor{color2}{rgb}{0.172549019607843,0.627450980392157,0.172549019607843}

\begin{axis}[
axis line style={white!80!black},
legend cell align={left},
legend style={fill opacity=0.8, draw opacity=1, text opacity=1, at={(0.7,0.18)}, anchor=north west, draw=white!80!black},
tick align=outside,
tick pos=left,
x grid style= {white!80!black},
xmajorticks=false,
xmin=-0.7, xmax=0.7,
xtick style={color=white!15!black},
xtick={0},
xticklabels={COMPREHENSIBILITY\_SCORE},
y grid style={white!80!black},
ylabel={Comprehensibility Score},
ymajorgrids,
ymajorticks=true,
ymin=0, ymax=21,
ytick style={color=white!15!black}
]
\path [draw=color0, semithick]
(axis cs:-0.380,10.25)%x = -0.396 -> -0.380
--(axis cs:-0.040,10.25) %x = -0.004 -> -0.040
--(axis cs:-0.040,18)
--(axis cs:-0.380,18)
--(axis cs:-0.380,10.25)
--cycle;
\path [draw=color1, semithick]
(axis cs:0.040,8)
--(axis cs:0.380,8)
--(axis cs:0.380,15)
--(axis cs:0.040,15)
--(axis cs:0.040,8)
--cycle;
\draw[draw=white!28.2352941176471!black,fill=color0,line width=0.3pt] (axis cs:0,0) rectangle (axis cs:0,0);
\addlegendimage{area legend,draw=color0,fill=white,line width=0.3pt};
\addlegendentry{SHAP}

\draw[draw=white!28.2352941176471!black,fill=white,line width=0.3pt] (axis cs:0,0) rectangle (axis cs:0,0);
\addlegendimage{area legend,draw=color1,fill=white,line width=0.3pt};
\addlegendentry{LIME}

\addplot [semithick, color0, forget plot]
table {%
-0.2 10.25
-0.2 1
};
\addplot [semithick, color0, forget plot]
table {%
-0.2 18
-0.2 20
};
\addplot [semithick, color0, forget plot]
table {%
-0.298 1
-0.102 1
};
\addplot [semithick, color0, forget plot]
table {%
-0.298 20
-0.102 20
};
\addplot [semithick, color1, forget plot]
table {%
0.2 8
0.2 1
};
\addplot [semithick, color1, forget plot]
table {%
0.2 15
0.2 20
};
\addplot [semithick, color1, forget plot]
table {%
0.102 1
0.298 1
};
\addplot [semithick, color1, forget plot]
table {%
0.102 20
0.298 20
};
\addplot [semithick, color0, forget plot]
table {%
-0.380 15
-0.040 15
};
\addplot [color0, mark=asterisk, mark size=3, mark options={solid}, only marks, forget plot]
table {%
-0.2 13.7272727272727
};
\addplot [semithick, color1, forget plot]
table {%
0.040 12.5
0.380 12.5
};
\addplot [color1, mark=asterisk, mark size=3, mark options={solid,fill=black}, only marks, forget plot]
table {%
0.2 11.4166666666667
};
\draw (axis cs:-0.2,15.28) node[
  scale=0.7,
  text=black,
  rotate=0.0
]{\bfseries 15.3};
\draw (axis cs:0.2,12.78) node[
  scale=0.7,
  text=black,
  rotate=0.0
]{\bfseries 12.8};
\end{axis}

\end{tikzpicture}
    \caption{LIME and SHAP's comprehensibility score. We included the mean values with stars, and included the median values above the median lines of each box-plot. No significant difference is visible between these two XAI methods.}
    \label{fig:RQ1_comprehensibility}
\end{figure}
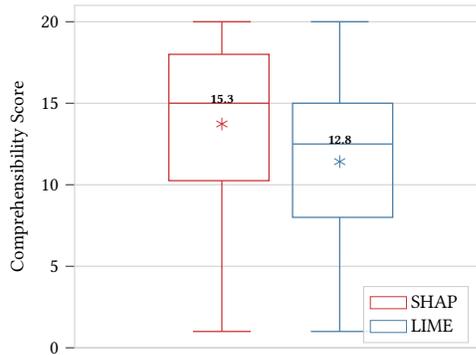

We collected responses from 46 participants, each answering the questions above for either LIME (24) or SHAP (22), and computed the overall comprehensibility score for each user. Figure~\ref{fig:RQ1_comprehensibility} shows the minimum, maximum, sample median, as well as the first and third quartiles of the comprehensibility scores for both methods. 
Post-hoc comparison of mean values using a two-sample t-test revealed that there is no significant difference (t=-1.56, p=0.12) between the mean comprehensibility score of SHAP (13.73) and LIME (11.42). This result shows that LIME and SHAP are equally comprehensible by the participants. 
%\textcolor{red}{We also conducted a Wilcoxon-Mann-Whitney test, which confirmed a significant difference between SHAP and LIME's score distribution (t-value=184.5, p=0.040).}

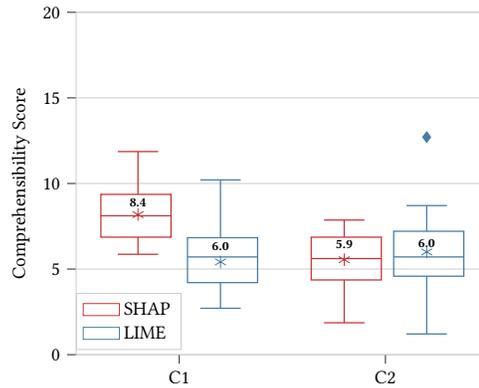
\begin{figure}[t]
    \centering
    % This file was created by tikzplotlib v0.9.7.
\begin{tikzpicture}[scale=0.8]

\definecolor{color0}{rgb}{0.795098039215686,0.200980392156863,0.206862745098039}
\definecolor{color1}{rgb}{0.278921568627451,0.487745098039216,0.658333333333333}
\definecolor{color2}{rgb}{0.172549019607843,0.627450980392157,0.172549019607843}

\begin{axis}[
axis line style={white!80!black},
legend cell align={left},
legend style={fill opacity=0.8, draw opacity=1, text opacity=1, at={(0,0.18)}, anchor=north west, draw=white!80!black},
tick align=outside,
tick pos=left,
x grid style={white!80!black},
xmajorticks=false,
xmin=-0.5, xmax=1.5,
xtick style={color=white!15!black},
xtick={0,1},
xticklabels={C1,C2},
xmajorticks=true,
y grid style={white!80!black},
ylabel={Comprehensibility Score},
ymajorgrids,
ymajorticks=true,
ymin=0, ymax=20,
ytick style={color=white!15!black}
]
\path [draw=color0, semithick]
(axis cs:-0.380,6.86363636363636) 
--(axis cs:-0.040,6.86363636363636) 
--(axis cs:-0.040,9.36363636363636)
--(axis cs:-0.380,9.36363636363636)
--(axis cs:-0.380,6.86363636363636)
--cycle;
\path [draw=color1, semithick]
(axis cs:0.040,4.20833333333333)
--(axis cs:0.380,4.20833333333333)
--(axis cs:0.380,6.83333333333333)
--(axis cs:0.040,6.83333333333333)
--(axis cs:0.040,4.20833333333333)
--cycle;
\path [draw=color0, semithick]
(axis cs:0.640,4.36363636363636)
--(axis cs:0.980,4.36363636363636)
--(axis cs:0.980,6.86363636363636)
--(axis cs:0.640,6.86363636363636)
--(axis cs:0.640,4.36363636363636)
--cycle;
\path [draw=color1, semithick]
(axis cs:1.040,4.58333333333333)  %x: 1.004 -> 1.040
--(axis cs:1.380,4.58333333333333) %x: 1.396 -> 1.380
--(axis cs:1.380,7.20833333333333)
--(axis cs:1.040,7.20833333333333)
--(axis cs:1.040,4.58333333333333)
--cycle;
\draw[draw=color0,fill=white,line width=0.3pt] (axis cs:0,0) rectangle (axis cs:0,0);
\addlegendimage{area legend,draw=color0,fill=white,line width=0.3pt};
\addlegendentry{SHAP}

\draw[draw=color1,fill=white,line width=0.3pt] (axis cs:0,0) rectangle (axis cs:0,0);
\addlegendimage{area legend,draw=color1,fill=white,line width=0.3pt};
\addlegendentry{LIME}

\addplot [semithick, color0, forget plot]
table {%
-0.2 6.86363636363636
-0.2 5.86363636363636
};
\addplot [semithick, color0, forget plot]
table {%
-0.2 9.36363636363636
-0.2 11.8636363636364
};
\addplot [semithick, color0, forget plot]
table {%
-0.298 5.86363636363636
-0.102 5.86363636363636
};
\addplot [semithick, color0, forget plot]
table {%
-0.298 11.8636363636364
-0.102 11.8636363636364
};
\addplot [semithick, color1, forget plot]
table {%
0.2 4.20833333333333
0.2 2.70833333333333
};
\addplot [semithick, color1, forget plot]
table {%
0.2 6.83333333333333
0.2 10.2083333333333
};
\addplot [semithick, color1, forget plot]
table {%
0.102 2.70833333333333
0.298 2.70833333333333
};
\addplot [semithick, color1, forget plot]
table {%
0.102 10.2083333333333
0.298 10.2083333333333
};
\addplot [color0, mark=diamond*, mark size=2.5, mark options={solid}, only marks, forget plot]
table {%
0.2 -1.29166666666667
};
\addplot [semithick, color0, forget plot]
table {%
0.8 4.36363636363636
0.8 1.86363636363636
};
\addplot [semithick, color0, forget plot]
table {%
0.8 6.86363636363636
0.8 7.86363636363636
};
\addplot [semithick, color0, forget plot]
table {%
0.702 1.86363636363636
0.898 1.86363636363636
};
\addplot [semithick, color0, forget plot]
table {%
0.702 7.86363636363636
0.898 7.86363636363636
};
\addplot [semithick, color1, forget plot]
table {%
1.2 4.58333333333333
1.2 1.20833333333333
};
\addplot [semithick, color1, forget plot]
table {%
1.2 7.20833333333333
1.2 8.70833333333333
};
\addplot [semithick, color1, forget plot]
table {%
1.102 1.20833333333333
1.298 1.20833333333333
};
\addplot [semithick, color1, forget plot]
table {%
1.102 8.70833333333333
1.298 8.70833333333333
};
\addplot [color1, mark=diamond*, mark size=2.5, mark options={solid,fill=color1}, only marks, forget plot]
table {%
1.2 12.7083333333333
};
\addplot [semithick, color0, forget plot]
table {%
-0.380 8.11363636363636
-0.040 8.11363636363636
};
\addplot [color0, mark=asterisk, mark size=3, mark options={solid}, only marks, forget plot]
table {%
-0.2 8.18181818181818
};
\addplot [semithick, color1, forget plot]
table {%
0.040 5.70833333333333
0.380 5.70833333333333
};
\addplot [color1, mark=asterisk, mark size=3, mark options={solid}, only marks, forget plot]
table {%
0.2 5.41666666666666
};
\addplot [semithick, color0, forget plot]
table {%
0.640 5.61363636363636
0.980 5.61363636363636
};
\addplot [color0, mark=asterisk, mark size=3, mark options={solid}, only marks, forget plot]
table {%
0.8 5.54545454545454
};
\addplot [semithick, color1, forget plot]
table {%
1.040 5.70833333333333
1.380 5.70833333333333
};
\addplot [color1, mark=asterisk, mark size=3, mark options={solid}, only marks, forget plot]
table {%
1.2 6
};
\draw (axis cs:-0.2,8.89363636363636) node[
  scale=0.7,
  text=black,
  rotate=0.0
]{\bfseries 8.4};
\draw (axis cs:0.2,6.35833333333333) node[
  scale=0.7,
  text=black,
  rotate=0.0
]{\bfseries 6.0};
\draw (axis cs:0.8,6.4363636363636) node[
  scale=0.7,
  text=black,
  rotate=0.0
]{\bfseries 5.9};
\draw (axis cs:1.2,6.5) node[
  scale=0.7,
  text=black,
  rotate=0.0
]{\bfseries 6.0};
\end{axis}

\end{tikzpicture}
    \caption{C1 and C2 scores for both LIME and SHAP. For LIME, we observe no significant difference between C1 and C2 scores. On the other hand, we observe a significant decrease from C1 to C2 for SHAP's comprehensibility tasks.}
    \label{fig:RQ1_H1_C1_vs_C2}
\end{figure}

Next, we tested whether a model's confidence in the prediction, which we can measure by considering the PDD between possible classes, affects the users' comprehensibility. Recall that C1 is further away from the model's decision boundary than C2. Since the responses to these tasks represent variables from repeated measures groups, we follow~\cite{field2012discovering} and compare means by first calculating an adjustment factor for each user. We computed that factor by subtracting the participant's means (pMean) from the mean of both C1 and C2. We then add these adjustment factors to our participants' actual comprehensibility scores, comparing C1 and C2. Again, we did not see a significant difference (t= -0.80, p=0.42) between the mean comprehensibility scores of C1 (5.42) and C2 (6.0). However, as shown in Figure~\ref{fig:RQ1_H1_C1_vs_C2}, for SHAP, we measured a significant decrease (t= 5.54, p=0.00) between the scores of C1 (8.18) and C2 (5.55). These results show that SHAP's explainability visualizations were less comprehensible to the participants when the model's prediction was closer to the decision boundary. Moreover, we see a significant difference (t=4.51, p=0.00) between the mean comprehensibility score of SHAP (8.18) and LIME (5.42) for C1. That difference indicates that SHAP's users' comprehensibility was higher than LIME's for C1. The low score can be explained by the ``I do not know" answers on the F3 feature contribution.

% Qualitative analysis
In our first intermediate analysis, we noticed the difference between SHAP and LIME's comprehensibility scores, and therefore, to obtain further qualitative insight into the effect of the decision boundary distance on the participants' comprehensibility, we started asking them the following question right after the first assignment was over:

\begin{quote}
    Q1. Which task did you find more difficult (between C1 and C2)? And why?
\end{quote}

We collected the feedback from 20 participants and categorized their answers into three groups: i) ``C1 more difficult than C2'', ii) ``C2 more difficult than C1'', and iii) ``C1 and C2 similarly difficult''. %The details of our coding rules and keywords are presented in Appendix \ref{Appx_comprehensibility}.

Fifteen participants stated that they found answering the questions for C1, which has a higher distance from the decision boundary than C2, more complicated than C2. One participant, for instance, stated ``\emph{I did not know how to answer the questions and work with visualizations for C1. But for C2, it was clear for me how to use the visualizations and find my answers}''. This result indicates a learning effect involved and that participants became familiar with the visualizations over time while answering these samples' questions. We compared their feedback with their scores and noticed that none of these 15 participants answered the questions for C2 correctly, which indicates a mismatch between perceived and actual comprehensibility.
% !TeX root = ../AAAI-2022.tex

\section{RQ1: Predictability}
\label{RQ1_H2}

% Goal
After we reported on the \emph{comprehensibility} aspect of the first research question, RQ1, in the previous section, we now focus on the \emph{predictability} aspect.

\begin{figure}
  \centering
  \input{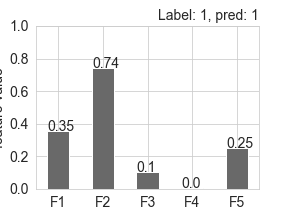}
  \caption{Example SHAP explanation (EX05) shown to the user as part of Assignment 2. The sample is classified as medium-price (label 1) and depicted as bar plot with its actual label and the model's prediction next its plot.}
  \label{fig:RQ1_trust_SHAP}
  %\vspace{-0.5cm}
\end{figure}

% Method
For that purpose, we presented the second assignment to the participants (see Figure~\ref{fig:survey_flow}) to analyze whether they can predict the model's behavior and detect the misclassification using the information they receive from the explanations. Figure~\ref{fig:RQ1_trust_SHAP} shows one of the three SHAP explanations we presented to the user. In contrast to the previous assignment, each was classified differently as low-, medium-, or high-price. In two tasks, we also introduced misclassified samples from medium- to low-price (T1) and from high- to medium-price (T2).

We asked the participants to answer the following multiple-choice questions for the test samples in each task:

\begin{enumerate}
    \label{trust_questions}
    \item What would the model predict when considering only the value of F1. 
    \item What would the model predict when we consider the value of F2 and F1?
    \item What is the effect on class probabilities if we consider F1, F2, and F3 (or F5) values?
    \item What does the model predict for this sample?
\end{enumerate}

The fourth question is about the model's behavior and follows a different scoring mechanism than the other questions. Table~\ref{table:RQ1_trust} lists the possible answers and the scoring rules we applied. Following our comprehensibility scoring, we give each correct answer (yes) a score of 2, and each wrong answer (no) a score of -1. For example, a participant that guesses the label correctly but predicts the model prediction wrong, receives a score of 1 (2 + -1).   However, participants who suspect the misclassification receive a score of 1 for correctly guessing the model's failure.

\begin{table}[t]
\centering
  \caption{Scoring rules for the last question in Assignment 2. The rules are based on what the user detected.}
  \label{table:RQ1_trust}
  %\scalebox{1}{
  \begin{tabular}{ccc}
    \toprule
     score & detected correct & detected prediction \\
           & label            & correctly \\
    \midrule
    
     4 & yes & yes \\ % e.g. if user.predict(label(x)) == label(x) to user.predict(model.predict(x)) == model.predict(x)
    \midrule
    
     1 & no & yes \\ % e.g. if user.predict(label(x)) != label(x) to user.predict(model.predict(x)) == model.predict(x)
    \midrule
    
     1 & yes & no \\ % e.g. if user.predict(label(x)) == label(x) to user.predict(model.predict(x)) != model.predict(x)
    \midrule
    
     -2 & no & no \\ % e.g. if user.predict(label(x)) != label(x) to user.predict(model.predict(x)) != model.predict(x)
    \midrule
    
     1 & \multicolumn{2}{c}{User suspects the misclassification} \\
       & \multicolumn{2}{c}{when the model misclassifies a sample}\\
    \midrule
      
    0 & \multicolumn{2}{c}{User answers with ``I do not know''} \\
  \bottomrule
\end{tabular}
 %} % close rescale/resize environment
 %\vspace{-0.5cm}
\end{table}

\paragraph{Results}

Analogous to the comprehensibility score reported in the previous section, we computed a predictability score for both XAI methods. As shown in Figure~\ref{fig:RQ2_vs_RQ1} (RQ1), and confirmed by a two-sample t-test (t-value=-0.553, p=0.582) we found no significant difference between the mean predictability scores of SHAP (5.64) and LIME (6.38), which were both drawn from Gaussian distributions that were not significantly different from each other.

%\begin{figure}
%    \centering 
%    \input{figures/RQ1_H2}
%    \caption{LIME and SHAP's trust scores. No significant difference is visible between %these two XAI methods.}
%    \label{fig:RQ1_trust}
%\end{figure}

%% Categorize user answers
We further apply the Mann-Whitney test to analyze the difference between LIME and SHAP answer category distribution. We categorize the answers into three groups; first, the category of correct guesses on model prediction, second are the incorrect prediction guesses, and third is the neutral category "I do not know".
For SHAP with 22 participants and their answers for two tasks, this categorization results in 15 counts for the first category, 20 counts for the second category and 9 counts for neutral category. Categorization of LIME with 24 participants results in 20 counts for the first category, 23 counts for the second category and 5 counts for neutral category.
We find no significant difference between LIME and SHAP answer category distribution for this assignment (t-value=922.5, p=0.128).

% Qualitative Analysis
We also analyzed the participant's answers to see whether they could predict the model's behavior using this set of explained samples. We compared the first question of samples T1 and T2, which only considers the values of feature F1. We noticed that participants using LIME answered correctly more often than those using SHAP and that SHAP users struggled to find a threshold to decide whether the feature value increased the classification probability. LIME explanations, on the other hand, provide such a threshold range.

For the second question, we noticed that participants using SHAP scored better than those using LIME. Based on LIME participants' feedback, the effect of F1 and F2 values on medium-price class was unclear to them.  

As part of the third question, we asked the participants about the impact of feature value F3. We wanted to know whether it pushes a decision towards a class and away from another class. The average scores of both participants groups, SHAP and LIME, were low for this question, with a mean score of 0.27 and 0.66, respectively. Fourteen participants using SHAP answered wrong, and 4 participants answered with ``I do not know''. On the other hand, 8 Participants using LIME answered wrong, and 6 participants answered with ``I do not know''.

Finally, the participants performed equally well when answering the fourth question for SHAP and LIME. Breaking down the scores and looking deeper, we noticed that SHAP participants often predicted the label of a sample correctly (15 from 22 participants) but failed to predict the model's behavior. On the other hand, LIME participants detected model misclassifications more often and predicted labels correctly (15 out of 24 participants).
% !TeX root = ../AAAI-2022.tex

\section{RQ2: Improving Predictability with Visualizations}
\label{RQ2}

% Goal
We now address our second research question, RQ2, and examine how visualizations of misclassified and counterfactual samples can improve the users' predictability.

% Method
%For that purpose, Assignment 3 uses the same set of questions as Assignment 2 but presents additional explanations about the misclassified samples to the user. This selection allows us to compare trust scores across assignments and measure how they are affected by additional visualizations.

%We also defined two additional tasks (T5 and T6) with two additional samples, which Figure~\ref{fig:survey_flow} illustrates. The sample presented in T5 is the same as in T3 (medium-price) but was misclassified as low-price by the model. For T6, the sample is the same as in T4 (high-price) but misclassified as medium-price. All chosen samples have low PDD values and are therefore close to the model's decision boundary. 

For that purpose, Assignment 3 uses the same set of questions and test samples as Assignment 2 but presents explanations about the misclassified and correctly classified samples to the user. This experiment allows us to compare predictability scores across assignments and measure how they are affected by additional visualizations. We illustrate the tasks and the assignment in Figure~\ref{fig:survey_flow}. All chosen samples have low PDD values and are close to the model's decision boundary. 

\paragraph{Results}

\begin{figure}
    \centering 
    % This file was created by tikzplotlib v0.9.7.
%\resizebox{0}{}{}{}
\begin{tikzpicture}[scale=0.8]

\definecolor{color0}{rgb}{0.795098039215686,0.200980392156863,0.206862745098039}
\definecolor{color1}{rgb}{0.278921568627451,0.487745098039216,0.658333333333333}
\definecolor{color2}{rgb}{0.172549019607843,0.627450980392157,0.172549019607843}

\begin{axis}[
axis line style={white!80!black},
legend cell align={left},
legend style={fill opacity=0.8, draw opacity=1, text opacity=1, at={(0,1)}, anchor=north west, draw=white!80!black},
tick align=outside,
tick pos=left,
x grid style={white!80!black},
xmajorticks=false,
xmin=-0.5, xmax=1.5,
xtick style={color=white!15!black},
xtick={0,1},
xticklabels={RQ1,RQ2},
xmajorticks=true,
y grid style={white!80!black},
ylabel={Predictability Score},
ymajorgrids,
ymajorticks=true,
ymin=-5, ymax=23,
ytick style={color=white!15!black}
]
\path [draw=color0, semithick]
(axis cs:-0.380,3)
--(axis cs:-0.040,3)
--(axis cs:-0.040,8.75)
--(axis cs:-0.380,8.75)
--(axis cs:-0.380,3)
--cycle;
\path [draw=color1, semithick]
(axis cs:0.040,3.75)
--(axis cs:0.380,3.75)
--(axis cs:0.380,8)
--(axis cs:0.040,8)
--(axis cs:0.040,3.75)
--cycle;
\path [draw=color0, semithick]
(axis cs:0.604,5.5)
--(axis cs:0.980,5.5)
--(axis cs:0.980,12.5)
--(axis cs:0.604,12.5)
--(axis cs:0.604,5.5)
--cycle;
\path [draw=color1, semithick]
(axis cs:1.040,9.75)
--(axis cs:1.380,9.75)
--(axis cs:1.380,14.75)
--(axis cs:1.040,14.75)
--(axis cs:1.040,9.75)
--cycle;
\draw[draw=color0,fill=white,line width=0.3pt] (axis cs:0,0) rectangle (axis cs:0,0);
\addlegendimage{area legend,draw=color0,fill=white,line width=0.3pt};
\addlegendentry{SHAP}

\draw[draw=color1,fill=white,line width=0.3pt] (axis cs:0,0) rectangle (axis cs:0,0);
\addlegendimage{area legend,draw=color1,fill=white,line width=0.3pt};
\addlegendentry{LIME}

\addplot [semithick, color0, forget plot]
table {%
-0.2 3
-0.2 -4
};
\addplot [semithick, color0, forget plot]
table {%
-0.2 8.75
-0.2 14
};
\addplot [semithick, color0, forget plot]
table {%
-0.298 -4
-0.102 -4
};
\addplot [semithick, color0, forget plot]
table {%
-0.298 14
-0.102 14
};
\addplot [semithick, color1, forget plot]
table {%
0.2 3.75
0.2 -2
};
\addplot [semithick, color1, forget plot]
table {%
0.2 8
0.2 14
};
\addplot [semithick, color1, forget plot]
table {%
0.102 -2
0.298 -2
};
\addplot [semithick, color1, forget plot]
table {%
0.102 14
0.298 14
};
\addplot [color1, mark=diamond*, mark size=2.5, mark options={solid}, only marks, forget plot]
table {%
0.2 17
};
\addplot [semithick, color0, forget plot]
table {%
0.8 5.5
0.8 -1
};
\addplot [semithick, color0, forget plot]
table {%
0.8 12.5
0.8 17
};
\addplot [semithick, color0, forget plot]
table {%
0.702 -1
0.898 -1
};
\addplot [semithick, color0, forget plot]
table {%
0.702 17
0.898 17
};
\addplot [semithick, color1, forget plot]
table {%
1.2 9.75
1.2 4
};
\addplot [semithick, color1, forget plot]
table {%
1.2 14.75
1.2 20
};
\addplot [semithick, color1, forget plot]
table {%
1.102 4
1.298 4
};
\addplot [semithick, color1, forget plot]
table {%
1.102 20
1.298 20
};
\addplot [semithick, color0, forget plot]
table {%
-0.380 6
-0.040 6
};
\addplot [color0, mark=asterisk, mark size=3, mark options={solid}, only marks, forget plot]
table {%
-0.2 5.63636363636364
};
\addplot [semithick, color1, forget plot]
table {%
0.040 5.5
0.380 5.5
};
\addplot [color1, mark=asterisk, mark size=3, mark options={solid}, only marks, forget plot]
table {%
0.2 6.375
};
\addplot [semithick, color0, forget plot]
table {%
0.604 8
0.980 8
};
\addplot [color0, mark=asterisk, mark size=3, mark options={solid}, only marks, forget plot]
table {%
0.8 8.86363636363636
};
\addplot [semithick, color1, forget plot]
table {%
1.040 11
1.380 11
};
\addplot [color1, mark=asterisk, mark size=3, mark options={solid}, only marks, forget plot]
table {%
1.2 11.7916666666667
};
\draw (axis cs:-0.2,7) node[
  scale=0.7,
  text=black,
  rotate=0.0
]{\bfseries 6.3};
\draw (axis cs:0.2,5) node[
  scale=0.7,
  text=black,
  rotate=0.0
]{\bfseries 5.8};
\draw (axis cs:0.8,7) node[
  scale=0.7,
  text=black,
  rotate=0.0
]{\bfseries 8.3};
\draw (axis cs:1.2,10.5) node[
  scale=0.7,
  text=black,
  rotate=0.0
]{\bfseries 11.3};
\end{axis}

\end{tikzpicture}
    \caption{LIME's and SHAP's predictability score before and after the users study the explained counter-factual and misclassified samples. LIME shows significant improvement.}
    \label{fig:RQ2_vs_RQ1}
    %\vspace{-0.5cm}
\end{figure}
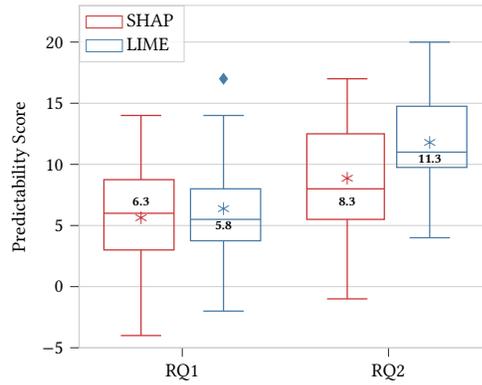

After checking that sample scores for both methods follow a Gaussian distribution, we compared the mean scores of both methods using a two-sample t-test and found that LIME's predictability score was significantly higher (t-value=-2.263, p=0.029) than that of SHAP. This result is also visible in Figure \ref{fig:RQ2_vs_RQ1}, which also shows that LIME's visualization of misclassified samples has a more substantial effect on improving the explainer's predictability.

%\begin{figure}
%    \centering 
%    \input{figures/RQ2_score}
%    \caption{LIME and SHAP's trust score after users studied explained counter-factual and misclassified samples. LIME participants achieved significantly higher score than SHAP participants.}
%    \label{fig:RQ2_score}
%\end{figure}

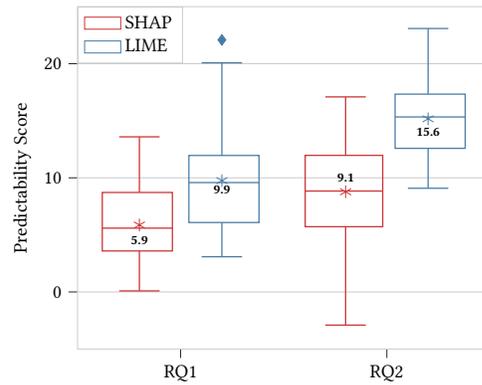
\begin{figure}
    \centering 
    % This file was created by tikzplotlib v0.9.7.
%\resizebox{0}{}{}{}
\begin{tikzpicture}[scale=0.8]

\definecolor{color0}{rgb}{0.795098039215686,0.200980392156863,0.206862745098039}
\definecolor{color1}{rgb}{0.278921568627451,0.487745098039216,0.658333333333333}
\definecolor{color2}{rgb}{0.172549019607843,0.627450980392157,0.172549019607843}

\begin{axis}[
axis line style={white!80!black},
legend cell align={left},
legend style={fill opacity=0.8, draw opacity=1, text opacity=1, at={(0,1)}, anchor=north west, draw=white!80!black},
tick align=outside,
tick pos=left,
x grid style={white!80!black},
xmajorticks=false,
xmin=-0.5, xmax=1.5,
xtick style={color=white!15!black},
xtick={0,1},
xticklabels={RQ1,RQ2},
xmajorticks=true,
y grid style={white!80!black},
ylabel={Predictability Score},
ymajorgrids,
ymajorticks=true,
ymin=-5, ymax=25,
ytick style={color=white!15!black}
]
\path [draw=color0, semithick]
(axis cs:-0.380,3.59090909090909)
--(axis cs:-0.040,3.59090909090909)
--(axis cs:-0.040,8.71590909090909)
--(axis cs:-0.380,8.71590909090909)
--(axis cs:-0.380,3.59090909090909)
--cycle;
\path [draw=color1, semithick]
(axis cs:0.040,6.08333333333333)
--(axis cs:0.380,6.08333333333333)
--(axis cs:0.380,11.9583333333333)
--(axis cs:0.040,11.9583333333333)
--(axis cs:0.040,6.08333333333333)
--cycle;
\path [draw=color0, semithick]
(axis cs:0.604,5.71590909090909)
--(axis cs:0.980,5.71590909090909)
--(axis cs:0.980,11.9659090909091)
--(axis cs:0.604,11.9659090909091)
--(axis cs:0.604,5.71590909090909)
--cycle;
\path [draw=color1, semithick]
(axis cs:1.040,12.5833333333333)
--(axis cs:1.380,12.5833333333333)
--(axis cs:1.380,17.3333333333333)
--(axis cs:1.040,17.3333333333333)
--(axis cs:1.040,12.5833333333333)
--cycle;
\draw[draw=color0,fill=white,line width=0.3pt] (axis cs:0,0) rectangle (axis cs:0,0);
\addlegendimage{area legend,draw=color0,fill=white,line width=0.3pt};
\addlegendentry{SHAP}

\draw[draw=color1,fill=white,line width=0.3pt] (axis cs:0,0) rectangle (axis cs:0,0);
\addlegendimage{area legend,draw=color1,fill=white,line width=0.3pt};
\addlegendentry{LIME}

\addplot [semithick, color0, forget plot]
table {%
-0.2 3.59090909090909
-0.2 0.0909090909090908
};
\addplot [semithick, color0, forget plot]
table {%
-0.2 8.71590909090909
-0.2 13.5909090909091
};
\addplot [semithick, color0, forget plot]
table {%
-0.298 0.0909090909090908
-0.102 0.0909090909090908
};
\addplot [semithick, color0, forget plot]
table {%
-0.298 13.5909090909091
-0.102 13.5909090909091
};
\addplot [semithick, color1, forget plot]
table {%
0.2 6.08333333333333
0.2 3.08333333333333
};
\addplot [semithick, color1, forget plot]
table {%
0.2 11.9583333333333
0.2 20.0833333333333
};
\addplot [semithick, color1, forget plot]
table {%
0.102 3.08333333333333
0.298 3.08333333333333
};
\addplot [semithick, color1, forget plot]
table {%
0.102 20.0833333333333
0.298 20.0833333333333
};
\addplot [color1, mark=diamond*, mark size=2.5, mark options={solid}, only marks, forget plot]
table {%
0.2 22.0833333333333
};
\addplot [semithick, color0, forget plot]
table {%
0.8 5.71590909090909
0.8 -2.90909090909091
};
\addplot [semithick, color0, forget plot]
table {%
0.8 11.9659090909091
0.8 17.0909090909091
};
\addplot [semithick, color0, forget plot]
table {%
0.702 -2.90909090909091
0.898 -2.90909090909091
};
\addplot [semithick, color0, forget plot]
table {%
0.702 17.0909090909091
0.898 17.0909090909091
};
\addplot [semithick, color1, forget plot]
table {%
1.2 12.5833333333333
1.2 9.08333333333333
};
\addplot [semithick, color1, forget plot]
table {%
1.2 17.3333333333333
1.2 23.0833333333333
};
\addplot [semithick, color1, forget plot]
table {%
1.102 9.08333333333333
1.298 9.08333333333333
};
\addplot [semithick, color1, forget plot]
table {%
1.102 23.0833333333333
1.298 23.0833333333333
};
\addplot [semithick, color0, forget plot]
table {%
-0.380 5.59090909090909
-0.040 5.59090909090909
};
\addplot [color0, mark=asterisk, mark size=3, mark options={solid}, only marks, forget plot]
table {%
-0.2 5.86363636363637
};
\addplot [semithick, color1, forget plot]
table {%
0.040 9.58333333333333
0.380 9.58333333333333
};
\addplot [color1, mark=asterisk, mark size=3, mark options={solid}, only marks, forget plot]
table {%
0.2 9.75000000000001
};
\addplot [semithick, color0, forget plot]
table {%
0.604 8.84090909090909
0.980 8.84090909090909
};
\addplot [color0, mark=asterisk, mark size=3, mark options={solid}, only marks, forget plot]
table {%
0.8 8.77272727272727
};
\addplot [semithick, color1, forget plot]
table {%
1.040 15.3333333333333
1.380 15.3333333333333
};
\addplot [color1, mark=asterisk, mark size=3, mark options={solid}, only marks, forget plot]
table {%
1.2 15.1666666666667
};
\draw (axis cs:-0.2,4.5) node[
  scale=0.7,
  text=black,
  rotate=0.0
]{\bfseries 5.9};
\draw (axis cs:0.2,9) node[
  scale=0.7,
  text=black,
  rotate=0.0
]{\bfseries 9.9};
\draw (axis cs:0.8,10) node[
  scale=0.7,
  text=black,
  rotate=0.0
]{\bfseries 9.1};
\draw (axis cs:1.2,14) node[
  scale=0.7,
  text=black,
  rotate=0.0
]{\bfseries 15.6};
\end{axis}

\end{tikzpicture}
    \caption{LIME's and SHAP's calculated adjusted mean of predictability score before and after each user studies the explained counter-factual and misclassified samples. Both methods show significant improvements.}
    \label{fig:RQ2_vs_RQ1_repeated_measure}
    %\vspace{-0.5cm}
\end{figure}

We further compared the total sum of the achieved predictability scores of Assignment 3 with those of the previously conducted Assignment 2 and see, as shown in Figure~\ref{fig:RQ2_vs_RQ1}, significant improvements for LIME (t-value=-4.387, p=0.0), but not for SHAP (t-value=-1.97, p=0.055). 
Since the responses in the Assignments 2 and 3 also represent variables from repeated measures groups, analogous to the comprehensibility score in RQ1, we follow Field et al.,~\cite{field2012discovering} method of comparing means for repeated measures by first calculating an adjustment factor for each user and adding this factor to their predictability scores. As shown in Figure~\ref{fig:RQ2_vs_RQ1_repeated_measure}, we see a significant improvement of the participants for both LIME (t-value=-4.301, p=0.0) and SHAP (t-value=-2.245, p=0.030). This result indicates that presenting explanations around the model's decision boundary to the user helps the user to understand the model's decision-making.

%% Categorize user answers
Analogous to our analysis in predictability RQ1, we apply the Mann-Whitney test to analyze the difference between LIME and SHAP category of answers distribution and find no significant difference (t-value=961.0, p=0.203). 
The categorization of SHAP answers results in 23 counts for the first category, 17 counts for the second category and 4 counts for neutral category. Categorization of LIME results in 31 counts for the first category, 15 counts for the second category and 2 counts for neutral category.
We further compare the category distribution between LIME answers in RQ1 and RQ2 and notice the distributions are significantly different (t-value=702.0, p=0.0). We find the same when comparing SHAP's answer category distribution between RQ1 and RQ2 (t-value=685.5, 0.004).

%Moreover, we observed improved answers for the third and fourth questions of these tasks, mentioned in~\ref{trust_questions}, which indicates that they could identify the features with lower contributions to the model's decision. They also tended to identify sample labels correctly and could predict the correct class for the misclassified test sample. From only 4 participants using LIME and 7 participants using SHAP, who answered these questions correctly for the second assignment, we see an improvement to 15 participants using LIME and 12 participants using SHAP, who answered the third assignment correctly. 

% Reduced text
Moreover, we observed that participants correctly identified sample labels and predicted the correct class for the misclassified test sample. From only 4 participants using LIME and 7 participants using SHAP, we see an improvement to 15 participants using LIME and 12 participants using SHAP, who answered the third assignment correctly.
% !TeX root = ../AAAI-2022.tex

\section{RQ3: Qualitative Analysis}
\label{RQ3}

% Goal
In this section, we answer our third research question, RQ3, studying the participants' feedback to provide a guideline for improving the design of XAI methods with local explanations. 

%Method
We seek to answer this question by qualitatively analyzing the participants' input. Therefore, at the end of the third assignment, we presented the participants with the following question:

\begin{quote}
    Q2. How confident were you when answering the questions? When did you feel less confident? 
\end{quote}

We followed Mayring's qualitative coding rules\cite{mayring2004qualitative} and coded the participants' responses into the following categories: i) high, ii) average, and iii) low self-confidence. This categorization resulted in balanced classes: 17 participants had high self-confidence using LIME (8) and SHAP (9). 13 participants had average self-confidence using LIME (6) and SHAP (7), and 10 participants had low self-confidence using LIME (10) and SHAP (6).

We further compared the participants' scores with their self-confidence category; we found that SHAP's participants' confidence category matches with their scores, indicating that participants who had high self-confidence using SHAP also correctly answered the assignments. 

However, we observed no correlation between the participants' self-confidence and the LIME scores. The median comprehensibility score (median score = 14.0) for 7 participants with average self-confidence was higher than both high-confidence (median score = 10.0) and low-confidence (median score = 11.0). LIME's predictability scores for both assignments two and three stayed the same and did not decrease with the participants' confidence category.

We compared the relationship between the participants' expertise (years of experience in ML and data science) with their scores for each category. We noticed a positive correlation for participants using SHAP, indicating more experienced participants could better interpret the explanations than participants with lower expertise. 

However, participants who used LIME and had low expertise achieved higher scores than the more experienced users. We considered the time participants needed to answer the questions and noticed that those who took more time achieved a higher score. The participants with high expertise and higher confidence often took less time to answer LIME's questions and had lower scores than the other user's who needed longer to answer the questions. 
% We can assume that participants with high expertise and high self-confidence answered intricate (sloppy) and did not take time to correct their interpretations. 

We noticed that participants using LIME agreed that misclassification information increased their confidence in their answers. However, participants using SHAP stated that the visualizations did not improve their confidence, and 6 participants stated that the misclassification information provided by SHAP did not help them at all and confused them more. Their feedback also correlates with their scores.

We continued our qualitative analysis and  presented the participants with the following questions:

\begin{quote}
    Q3. How much did the visualizations (of the second assignment) help to get an insight into the model and how it comes to its decisions?
\end{quote} 

We categorized the helpfulness and negative feedback of the participants into categories i) helpful to answer the test sample, ii) only helps to understand the explained samples, and iii) not helpful at all. We further clustered the negative feedback from the second and third categories to construct the improvement guidelines based on participants' needs. %For more detailed information on our coding, please refer to Appendix ~\ref{Appx_RQ3_feedback}.

Our coding resulted in having 36 of 46 participants (18 LIME and 17 SHAP) stating that the information was not enough to scale to a new sample. One participant who used LIME found the visualizations helpful to scale the information for a new sample. Ten participants (6 SHAP and 4 LIME) stated that they could not use the information and answered with very low confidence. We moved further to our fourth question to understand whether the misclassification and counter-factual samples help the participants gain more insight into the model's decision-making process.

\begin{quote}
    Q4. How much did the explanations of counter-factual and misclassified samples help to answer the questions of the third assignment?
\end{quote}

We cluster the participants' feedback based on their similarities to understand what confused them and caused them to fail in solving the assignments. %For more details on the coding rules, keywords, and the definition of each category, please refer to the Appendix \ref{Appx_RQ3_feedback}. 
We clustered their feedback into three categories; i) inconsistency of the explained values, ii) missing information from the visualizations, and iii) not at all helpful.

We first present the negative feedback from participants who used LIME; Two participants stated that assignments two and three were confusing. One had a low expertise rank, and the other had average expertise and no XAI experience. However, both scores significantly increased after the visualizations of misclassified samples and counter-factual explanations. One participant stated that \textit{"the way all the information was presented at once made it very difficult to understand the differences of the samples" (user\_id=27)}, and \textit{"I do not understand what the visualizations are trying to point out, they are all very similar to each other and identifying the correlation of the feature values on the target classes was not intuitive at all. I always used the bar plot and compared my sample with the labeled samples." (user\_id=39)}. 

Four LIME users were in the second\_category (missing information from the visualizations) and stated that they needed more explanations of more misclassified samples. They also mentioned that \textit{"the visualizations helped me understand why and when a misclassification might happen. Still, it was not enough to assume the model's behavior confidently." (user\_id=1)}. These participants also had significantly higher scores in the third assignment, indicating that the visualizations helped identify misclassifications regardless of their subjective feedback. 

Finally, only one LIME user stated that the information was completely confusing. the user could not make sense of the inequality range given by LIME's visualization and why some features had the same inequality range, even though they were classified into two different classes. When we explained that the inequalities depend on the feature interactions and why these changes stay the same for one feature, users admitted that they understood it. Still, it was not intuitive for the user at the assignment time. 

We conclude that the reason behind most users' confusion was the unexplained, inconsistent range of inequalities for samples from the same class. Moreover, LIME's local explanations only present that a feature value reduces the probability of the predicted class but does not reveal which class's probability increases. SHAP plots present this information.

We move on to negative feedback given by participants who used SHAP; No participant stated that SHAP's visualizations were not at all helpful. Only two stated that more explanations of misclassified samples would have helped them scale the explanations for a new sample ( category (ii)) and had a non-significant lower score t=1.0, p=0.5) for assignment three. Moreover, eight participants, who also achieved a higher score for the third assignment, mentioned that the visualizations confused them very much: \textit{"There were too many bars and numbers and finding the contribution of features to each class and the effect of feature values on other classes was very demanding and tiring" (user\_id=26)}. All these participants also stated that the inconsistency of the plot's scaling fooled them, and they had to invest more time to find the real contribution of feature values towards each class.  

We noticed that for SHAP, it still plots a long bar for each feature when all features have a low contribution. If the user does not look at the probability scales carefully, they might assume that all these features are equally contributing highly towards the respective class. 
% !TeX root = ../AAAI-2022.tex

\section{Discussion \& Design Recommendation}
\label{discussion}

We now summarize the key findings of our user study and propose a set of design recommendations that can substantially contribute to the design of new or refinement of existing XAI methods:

\begin{itemize}

    \item \emph{Explanations should be consistent}. This may seem obvious, but when evaluating LIME, we noticed that many participants were confused by inconsistent inequality ranges for the same feature in different samples of the same class (e.g., model predicted two instances as low-price because one instance had 0.26<F1-value<0.36, while another instance had F1-value<0.26). Such inconsistencies could be explained by showing the correlation between feature values. %(from section RQ3: Qualitative Analysis).
    
    \item \emph{Explanations should have fixed scales}. Participants working with SHAP tended to interpret feature explanations incorrectly for features with smaller scales. The visualizations ``fooled'' them, and they assumed that a feature with a higher bar has a greater influence on a sample, even though its SHAP value was much smaller than in another sample. %(from section RQ3: Qualitative Analysis).
    
    \item \emph{Explanations should also provide counterfactual examples}. Our quantitative and qualitative results show that participants predict model behavior better using the explanations if counterfactual samples are presented. These additional explanations could also help users understand the differences between the classes more clearly. %(from sections RQ2: Improving Trust with Visualizations and RQ3: Qualitative Analysis).
    
    \item \emph{Explanations should contain misclassified samples close to the decision boundary}. From our experiments, we learned that participants predict model's decisions with explanations better when they see misclassified examples and understand why the model made a wrong decision. This can be achieved by presenting samples close to a model's decision boundary, which of course, are often subject to misclassification. %(from sections RQ2: Improving Trust with Visualizations and RQ3: Qualitative Analysis).
    
    \item \emph{Explanations should contain correctly predicted samples close to the decision boundary}. Our results also indicate that participants achieved higher scores when being presented with explanations on correct predictions of samples. These samples can be identified by selecting correctly classified samples with low prediction confidence. %(from sections RQ2: Improving Trust with Visualizations and RQ3: Qualitative Analysis).
    
\end{itemize}

% Limitations
Our work is, of course, currently limited to comparing the local explanations of two XAI approaches, LIME and SHAP. Also, the Boston housing dataset we used in our experiments has been simplified to tabular data with a few dimensions. Moreover, users neither had the option to choose the samples themselves nor did we allow them to interact with the model and different outputs of the XAI approaches. However, we argue that these restrictions were necessary to reduce the number of confounding variables in our user study, which could influence our variables by events that are not causally related. We also believe that our experimental setup provides the necessary degree of generalizability to be transferred to the evaluation of other explanation tools.

% Potential future research
Potential future work could expand our approach to other types of data (e.g., acoustic or sensor data) and other emerging XAI techniques (e.g., LORE~\cite{LORE2018}), 
we have not yet considered in our explanations. Moreover, one could compare local and global explanation designs and study how they affect users' comprehensibility and predictability. Another potentially interesting research direction is to investigate how active learning techniques could be used to support users in comprehending and predicting model decisions~\cite{mondal2020alex}.

% Final, take-away statement
Overall, we believe that user studies should become an integral part of the improvement of existing and the development of new XAI methods. Since explanations must ultimately be understood by users, their perceptions and interpretations of explanations should also be systematically analyzed and understood. This can improve the XAI methods and also the user interaction with these methods.
% !TeX root = ../AAAI-2022.tex

\section{Conclusion}
\label{conclusion}

In this paper, we conducted a user study to investigate how well users comprehend the explanations and predict model behavior provided by two widely used tools, LIME, and SHAP.
We measured the comprehensibility and predictability participants gained after interpreting a given set of local explanations to increase the comprehensiveness of the captured information for a new un-explained sample.
We formed our first research question to measure comprehensibility and predictability. We showed that the comprehensibility of SHAP explanations significantly decreases for samples close to the decision boundary. 
Second, we studied the information participants require to gain a more global interpretation of the model behavior and to increase their predictability using explanations. We observed that explaining misclassified and counterfactual samples to the participants can significantly improve their predictability (especially with LIME explanations). They recognized the model behavior for unexplained samples close to the model decision boundary. 
Furthermore, our qualitative analysis of participants' feedback indicated that they require information such as justifying the explained values (LIME inequalities or SHAP values) to correctly interpret the model behavior and move towards a more global interpretation of the model decision boundary. 
Finally, we learned that the users' confidence in interpreting the explanations strongly relies on the diversity and quantity of the explained samples. The more different instances were studied by participants, the more accurately they could interpret the outputs of the explainability approaches and predict the decisions of the model.
\section{Acknowledgments}
%Acknowledgement is removed for a double-blind review.
We thank all the volunteers, and all the reviewers, who wrote and provided helpful comments on previous versions of this document. We specially thank our colleagues, Clemens Heistracher and Denis Katic for their constructive feedback on the structure of this work. 
%We further like to thank Dr. Philipp Wintersberger for his constructive feedback and insight to this work. 
We further like to thank Dr. Jasmin Lampert for her constructive feedback and insight to this work. 
We also thank the Austrian Research Promotion Agency (FFG) for funding this work, which is a part of the industrial project DeepRUL, project ID 871357.

\bibliographystyle{ACM-Reference-Format}
\bibliography{references}


\begin{thebibliography}{37}

%%% the Web address.

\ifx \showCODEN    \undefined \def \showCODEN     #1{\unskip}     \fi
\ifx \showDOI      \undefined \def \showDOI       #1{#1}\fi
\ifx \showISBNx    \undefined \def \showISBNx     #1{\unskip}     \fi
\ifx \showISBNxiii \undefined \def \showISBNxiii  #1{\unskip}     \fi
\ifx \showISSN     \undefined \def \showISSN      #1{\unskip}     \fi
\ifx \showLCCN     \undefined \def \showLCCN      #1{\unskip}     \fi
\ifx \shownote     \undefined \def \shownote      #1{#1}          \fi
\ifx \showarticletitle \undefined \def \showarticletitle #1{#1}   \fi
\ifx \showURL      \undefined \def \showURL       {\relax}        \fi
% invisible to TeX
\providecommand\bibfield[2]{#2}
\providecommand\bibinfo[2]{#2}
\providecommand\natexlab[1]{#1}
\providecommand\showeprint[2][]{arXiv:#2}

\bibitem[\protect\citeauthoryear{Alqaraawi, Schuessler, Wei\ss{}, Costanza, and
  Berthouze}{Alqaraawi et~al\mbox{.}}{2020}]%
        {alqaraawi2020evaluating}
\bibfield{author}{\bibinfo{person}{Ahmed Alqaraawi}, \bibinfo{person}{Martin
  Schuessler}, \bibinfo{person}{Philipp Wei\ss{}}, \bibinfo{person}{Enrico
  Costanza}, {and} \bibinfo{person}{Nadia Berthouze}.}
  \bibinfo{year}{2020}\natexlab{}.
\newblock \showarticletitle{Evaluating Saliency Map Explanations for
  Convolutional Neural Networks: A User Study}. In
  \bibinfo{booktitle}{\emph{Proceedings of the 25th International Conference on
  Intelligent User Interfaces}} (Cagliari, Italy) \emph{(\bibinfo{series}{IUI
  '20})}. \bibinfo{publisher}{Association for Computing Machinery},
  \bibinfo{address}{New York, NY, USA}, \bibinfo{pages}{275–285}.
\newblock
\showISBNx{9781450371186}
\urldef\tempurl%
\url{https://doi.org/10.1145/3377325.3377519}
\showDOI{\tempurl}


\bibitem[\protect\citeauthoryear{Carvalho, Pereira, and Cardoso}{Carvalho
  et~al\mbox{.}}{2019}]%
        {carvalho2019machine}
\bibfield{author}{\bibinfo{person}{Diogo~V Carvalho},
  \bibinfo{person}{Eduardo~M Pereira}, {and} \bibinfo{person}{Jaime~S
  Cardoso}.} \bibinfo{year}{2019}\natexlab{}.
\newblock \showarticletitle{Machine learning interpretability: A survey on
  methods and metrics}.
\newblock \bibinfo{journal}{\emph{Electronics}} \bibinfo{volume}{8},
  \bibinfo{number}{8} (\bibinfo{year}{2019}), \bibinfo{pages}{832}.
\newblock


\bibitem[\protect\citeauthoryear{Field, Miles, and Field}{Field
  et~al\mbox{.}}{2012}]%
        {field2012discovering}
\bibfield{author}{\bibinfo{person}{Andy~P Field}, \bibinfo{person}{Jeremy
  Miles}, {and} \bibinfo{person}{Zo{\"e} Field}.}
  \bibinfo{year}{2012}\natexlab{}.
\newblock \bibinfo{booktitle}{\emph{Discovering statistics using R}}.
\newblock \bibinfo{publisher}{SAGE publications}, \bibinfo{address}{London,
  England}. 361--365 pages.
\newblock
\showISBNx{9781446200469}


\bibitem[\protect\citeauthoryear{Guidotti, Monreale, Ruggieri, Pedreschi,
  Turini, and Giannotti}{Guidotti et~al\mbox{.}}{2018a}]%
        {LORE2018}
\bibfield{author}{\bibinfo{person}{Riccardo Guidotti}, \bibinfo{person}{Anna
  Monreale}, \bibinfo{person}{Salvatore Ruggieri}, \bibinfo{person}{Dino
  Pedreschi}, \bibinfo{person}{Franco Turini}, {and} \bibinfo{person}{Fosca
  Giannotti}.} \bibinfo{year}{2018}\natexlab{a}.
\newblock \showarticletitle{Local rule-based explanations of black box decision
  systems}.
\newblock \bibinfo{journal}{\emph{arXiv preprint arXiv:1805.10820}}
  (\bibinfo{year}{2018}).
\newblock


\bibitem[\protect\citeauthoryear{Guidotti, Monreale, Ruggieri, Turini,
  Giannotti, and Pedreschi}{Guidotti et~al\mbox{.}}{2018b}]%
        {guidotti2018survey}
\bibfield{author}{\bibinfo{person}{Riccardo Guidotti}, \bibinfo{person}{Anna
  Monreale}, \bibinfo{person}{Salvatore Ruggieri}, \bibinfo{person}{Franco
  Turini}, \bibinfo{person}{Fosca Giannotti}, {and} \bibinfo{person}{Dino
  Pedreschi}.} \bibinfo{year}{2018}\natexlab{b}.
\newblock \showarticletitle{A survey of methods for explaining black box
  models}.
\newblock \bibinfo{journal}{\emph{ACM computing surveys (CSUR)}}
  \bibinfo{volume}{51}, \bibinfo{number}{5} (\bibinfo{year}{2018}),
  \bibinfo{pages}{1--42}.
\newblock


\bibitem[\protect\citeauthoryear{Harrison~Jr and Rubinfeld}{Harrison~Jr and
  Rubinfeld}{1978}]%
        {harrison1978hedonic}
\bibfield{author}{\bibinfo{person}{David Harrison~Jr} {and}
  \bibinfo{person}{Daniel~L Rubinfeld}.} \bibinfo{year}{1978}\natexlab{}.
\newblock \showarticletitle{Hedonic housing prices and the demand for clean
  air}.
\newblock \bibinfo{journal}{\emph{Journal of environmental economics and
  management}} \bibinfo{volume}{5}, \bibinfo{number}{1} (\bibinfo{year}{1978}),
  \bibinfo{pages}{81--102}.
\newblock


\bibitem[\protect\citeauthoryear{Hart et~al\mbox{.}}{Hart
  et~al\mbox{.}}{1988}]%
        {hart1988development}
\bibfield{author}{\bibinfo{person}{SG Hart} {et~al\mbox{.}}}
  \bibinfo{year}{1988}\natexlab{}.
\newblock \bibinfo{title}{Development of NASA-TLX: Results of empirical and
  theoretical research.” inP. A. Hancock and N. Meshkati (eds.), Human Mental
  Workload}.
\newblock
\newblock


\bibitem[\protect\citeauthoryear{Hase and Bansal}{Hase and Bansal}{2020}]%
        {hase2020evaluating}
\bibfield{author}{\bibinfo{person}{Peter Hase} {and} \bibinfo{person}{Mohit
  Bansal}.} \bibinfo{year}{2020}\natexlab{}.
\newblock \showarticletitle{Evaluating Explainable AI: Which Algorithmic
  Explanations Help Users Predict Model Behavior?}
\newblock \bibinfo{journal}{\emph{Association for Computational Linguistics
  (ACL)}} (\bibinfo{year}{2020}).
\newblock


\bibitem[\protect\citeauthoryear{Haufe, Meinecke, G{\"o}rgen, D{\"a}hne,
  Haynes, Blankertz, and Bie{\ss}mann}{Haufe et~al\mbox{.}}{2014}]%
        {haufe2014interpretation}
\bibfield{author}{\bibinfo{person}{Stefan Haufe}, \bibinfo{person}{Frank
  Meinecke}, \bibinfo{person}{Kai G{\"o}rgen}, \bibinfo{person}{Sven
  D{\"a}hne}, \bibinfo{person}{John-Dylan Haynes}, \bibinfo{person}{Benjamin
  Blankertz}, {and} \bibinfo{person}{Felix Bie{\ss}mann}.}
  \bibinfo{year}{2014}\natexlab{}.
\newblock \showarticletitle{On the interpretation of weight vectors of linear
  models in multivariate neuroimaging}.
\newblock \bibinfo{journal}{\emph{Neuroimage}}  \bibinfo{volume}{87}
  (\bibinfo{year}{2014}), \bibinfo{pages}{96--110}.
\newblock


\bibitem[\protect\citeauthoryear{Hoffman}{Hoffman}{2017}]%
        {hoffman2017taxonomy}
\bibfield{author}{\bibinfo{person}{Robert~R Hoffman}.}
  \bibinfo{year}{2017}\natexlab{}.
\newblock \showarticletitle{A taxonomy of emergent trusting in the
  human--machine relationship}.
\newblock \bibinfo{journal}{\emph{Cognitive systems engineering: The future for
  a changing world}} (\bibinfo{year}{2017}), \bibinfo{pages}{137--164}.
\newblock


\bibitem[\protect\citeauthoryear{Jacovi, Marasovi{\'c}, Miller, and
  Goldberg}{Jacovi et~al\mbox{.}}{2021}]%
        {jacovi2021formalizing}
\bibfield{author}{\bibinfo{person}{Alon Jacovi}, \bibinfo{person}{Ana
  Marasovi{\'c}}, \bibinfo{person}{Tim Miller}, {and} \bibinfo{person}{Yoav
  Goldberg}.} \bibinfo{year}{2021}\natexlab{}.
\newblock \showarticletitle{Formalizing trust in artificial intelligence:
  Prerequisites, causes and goals of human trust in AI}. In
  \bibinfo{booktitle}{\emph{Proceedings of the 2021 ACM conference on fairness,
  accountability, and transparency}}. \bibinfo{pages}{624--635}.
\newblock


\bibitem[\protect\citeauthoryear{Kaur, Nori, Jenkins, Caruana, Wallach, and
  Wortman~Vaughan}{Kaur et~al\mbox{.}}{2020}]%
        {kaur2020interpreting}
\bibfield{author}{\bibinfo{person}{Harmanpreet Kaur}, \bibinfo{person}{Harsha
  Nori}, \bibinfo{person}{Samuel Jenkins}, \bibinfo{person}{Rich Caruana},
  \bibinfo{person}{Hanna Wallach}, {and} \bibinfo{person}{Jennifer
  Wortman~Vaughan}.} \bibinfo{year}{2020}\natexlab{}.
\newblock \showarticletitle{Interpreting Interpretability: Understanding Data
  Scientists' Use of Interpretability Tools for Machine Learning}. In
  \bibinfo{booktitle}{\emph{Proceedings of the 2020 CHI Conference on Human
  Factors in Computing Systems}} (Honolulu, HI, USA)
  \emph{(\bibinfo{series}{CHI '20})}. \bibinfo{publisher}{Association for
  Computing Machinery}, \bibinfo{address}{New York, NY, USA},
  \bibinfo{pages}{1–14}.
\newblock
\showISBNx{9781450367080}
\urldef\tempurl%
\url{https://doi.org/10.1145/3313831.3376219}
\showDOI{\tempurl}


\bibitem[\protect\citeauthoryear{Lakkaraju and Bastani}{Lakkaraju and
  Bastani}{2020}]%
        {lakkaraju2020fool}
\bibfield{author}{\bibinfo{person}{Himabindu Lakkaraju} {and}
  \bibinfo{person}{Osbert Bastani}.} \bibinfo{year}{2020}\natexlab{}.
\newblock \showarticletitle{"How Do I Fool You?": Manipulating User Trust via
  Misleading Black Box Explanations}. In \bibinfo{booktitle}{\emph{Proceedings
  of the AAAI/ACM Conference on AI, Ethics, and Society}} (New York, NY, USA)
  \emph{(\bibinfo{series}{AIES '20})}. \bibinfo{publisher}{Association for
  Computing Machinery}, \bibinfo{address}{New York, NY, USA},
  \bibinfo{pages}{79–85}.
\newblock
\showISBNx{9781450371100}
\urldef\tempurl%
\url{https://doi.org/10.1145/3375627.3375833}
\showDOI{\tempurl}


\bibitem[\protect\citeauthoryear{Lipton}{Lipton}{2018}]%
        {lipton2018mythos}
\bibfield{author}{\bibinfo{person}{Zachary~C Lipton}.}
  \bibinfo{year}{2018}\natexlab{}.
\newblock \showarticletitle{The Mythos of Model Interpretability: In machine
  learning, the concept of interpretability is both important and slippery.}
\newblock \bibinfo{journal}{\emph{Queue}} \bibinfo{volume}{16},
  \bibinfo{number}{3} (\bibinfo{year}{2018}), \bibinfo{pages}{31--57}.
\newblock


\bibitem[\protect\citeauthoryear{Lundberg and Lee}{Lundberg and Lee}{2017}]%
        {lundberg2017unified}
\bibfield{author}{\bibinfo{person}{Scott~M. Lundberg} {and}
  \bibinfo{person}{Su-In Lee}.} \bibinfo{year}{2017}\natexlab{}.
\newblock \showarticletitle{A Unified Approach to Interpreting Model
  Predictions}. In \bibinfo{booktitle}{\emph{Proceedings of the 31st
  International Conference on Neural Information Processing Systems}} (Long
  Beach, California, USA) \emph{(\bibinfo{series}{NIPS'17})}.
  \bibinfo{publisher}{Curran Associates Inc.}, \bibinfo{address}{Red Hook, NY,
  USA}, \bibinfo{pages}{4768–4777}.
\newblock
\showISBNx{9781510860964}


\bibitem[\protect\citeauthoryear{Martin}{Martin}{2007}]%
        {martin2007}
\bibfield{author}{\bibinfo{person}{David~W. Martin}.}
  \bibinfo{year}{2007}\natexlab{}.
\newblock \showarticletitle{Doing Psychology Experiments}.
\newblock  (\bibinfo{year}{2007}), \bibinfo{pages}{148--170}.
\newblock


\bibitem[\protect\citeauthoryear{Mayring}{Mayring}{2004}]%
        {mayring2004qualitative}
\bibfield{author}{\bibinfo{person}{Philipp Mayring}.}
  \bibinfo{year}{2004}\natexlab{}.
\newblock \showarticletitle{Qualitative content analysis}.
\newblock \bibinfo{journal}{\emph{A companion to qualitative research}}
  \bibinfo{volume}{1}, \bibinfo{number}{2} (\bibinfo{year}{2004}),
  \bibinfo{pages}{159--176}.
\newblock


\bibitem[\protect\citeauthoryear{Mohseni}{Mohseni}{2019}]%
        {mohseni2019toward}
\bibfield{author}{\bibinfo{person}{Sina Mohseni}.}
  \bibinfo{year}{2019}\natexlab{}.
\newblock \showarticletitle{Toward Design and Evaluation Framework for
  Interpretable Machine Learning Systems}. In
  \bibinfo{booktitle}{\emph{Proceedings of the 2019 AAAI/ACM Conference on AI,
  Ethics, and Society}}. \bibinfo{pages}{553--554}.
\newblock


\bibitem[\protect\citeauthoryear{Mohseni and Ragan}{Mohseni and Ragan}{2020}]%
        {mohseni2018human}
\bibfield{author}{\bibinfo{person}{Sina Mohseni} {and} \bibinfo{person}{Eric~D
  Ragan}.} \bibinfo{year}{2020}\natexlab{}.
\newblock \showarticletitle{A human-grounded evaluation benchmark for local
  explanations of machine learning}.
\newblock \bibinfo{journal}{\emph{arXiv preprint arXiv:1801.05075}}
  (\bibinfo{year}{2020}).
\newblock


\bibitem[\protect\citeauthoryear{Mohseni, Zarei, and Ragan}{Mohseni
  et~al\mbox{.}}{2021}]%
        {mohseni2021multidisciplinary}
\bibfield{author}{\bibinfo{person}{Sina Mohseni}, \bibinfo{person}{Niloofar
  Zarei}, {and} \bibinfo{person}{Eric~D Ragan}.}
  \bibinfo{year}{2021}\natexlab{}.
\newblock \showarticletitle{A multidisciplinary survey and framework for design
  and evaluation of explainable AI systems}.
\newblock \bibinfo{journal}{\emph{ACM Transactions on Interactive Intelligent
  Systems (TiiS)}} \bibinfo{volume}{11}, \bibinfo{number}{3-4}
  (\bibinfo{year}{2021}), \bibinfo{pages}{1--45}.
\newblock


\bibitem[\protect\citeauthoryear{Molnar, Casalicchio, and Bischl}{Molnar
  et~al\mbox{.}}{2020}]%
        {molnar2020interpretable}
\bibfield{author}{\bibinfo{person}{Christoph Molnar}, \bibinfo{person}{Giuseppe
  Casalicchio}, {and} \bibinfo{person}{Bernd Bischl}.}
  \bibinfo{year}{2020}\natexlab{}.
\newblock \showarticletitle{Interpretable Machine Learning--A Brief History,
  State-of-the-Art and Challenges}.
\newblock \bibinfo{journal}{\emph{Koprinska I. et al. (eds) ECML PKDD
  Workshops. ECML PKDD 2020. Communications in Computer and Information
  Science, vol 1323. Springer, Cham.}} (\bibinfo{year}{2020}).
\newblock


\bibitem[\protect\citeauthoryear{Mondal and Ganguly}{Mondal and
  Ganguly}{2020}]%
        {mondal2020alex}
\bibfield{author}{\bibinfo{person}{Ishani Mondal} {and}
  \bibinfo{person}{Debasis Ganguly}.} \bibinfo{year}{2020}\natexlab{}.
\newblock \showarticletitle{ALEX: Active Learning based Enhancement of a
  Classification Model's EXplainability}. In
  \bibinfo{booktitle}{\emph{Proceedings of the 29th ACM International
  Conference on Information \& Knowledge Management}}.
  \bibinfo{pages}{3309--3312}.
\newblock


\bibitem[\protect\citeauthoryear{Nourani, Kabir, Mohseni, and Ragan}{Nourani
  et~al\mbox{.}}{2019}]%
        {nourani2019effects}
\bibfield{author}{\bibinfo{person}{Mahsan Nourani}, \bibinfo{person}{Samia
  Kabir}, \bibinfo{person}{Sina Mohseni}, {and} \bibinfo{person}{Eric~D
  Ragan}.} \bibinfo{year}{2019}\natexlab{}.
\newblock \showarticletitle{The effects of meaningful and meaningless
  explanations on trust and perceived system accuracy in intelligent systems}.
  In \bibinfo{booktitle}{\emph{Proceedings of the AAAI Conference on Human
  Computation and Crowdsourcing}}, Vol.~\bibinfo{volume}{7}.
  \bibinfo{pages}{97--105}.
\newblock


\bibitem[\protect\citeauthoryear{Papenmeier, Englebienne, and
  Seifert}{Papenmeier et~al\mbox{.}}{2019}]%
        {papenmeier2019model}
\bibfield{author}{\bibinfo{person}{Andrea Papenmeier}, \bibinfo{person}{Gwenn
  Englebienne}, {and} \bibinfo{person}{Christin Seifert}.}
  \bibinfo{year}{2019}\natexlab{}.
\newblock \showarticletitle{How model accuracy and explanation fidelity
  influence user trust}.
\newblock \bibinfo{journal}{\emph{AI IJCAI Workshop on Explainable Artificial
  Intelligence}} (\bibinfo{year}{2019}).
\newblock


\bibitem[\protect\citeauthoryear{Qualtrics}{Qualtrics}{[n.d.]}]%
        {Qualtrics}
\bibfield{author}{\bibinfo{person}{Qualtrics}.}
  \bibinfo{year}{[n.d.]}\natexlab{}.
\newblock \bibinfo{booktitle}{\emph{Copyright Year: 2021, Location: Provo,
  Utah, USA}}.
\newblock
\urldef\tempurl%
\url{https://www.qualtrics.com}
\showURL{%
\tempurl}


\bibitem[\protect\citeauthoryear{Ribeiro, Singh, and Guestrin}{Ribeiro
  et~al\mbox{.}}{2016}]%
        {ribeiro2016should}
\bibfield{author}{\bibinfo{person}{Marco~Tulio Ribeiro},
  \bibinfo{person}{Sameer Singh}, {and} \bibinfo{person}{Carlos Guestrin}.}
  \bibinfo{year}{2016}\natexlab{}.
\newblock \showarticletitle{"Why should I trust you?" Explaining the
  predictions of any classifier}. In \bibinfo{booktitle}{\emph{Proceedings of
  the 22nd ACM SIGKDD International Conference on Knowledge Discovery and Data
  Mining}}. \bibinfo{pages}{1135--1144}.
\newblock


\bibitem[\protect\citeauthoryear{Ribeiro, Singh, and Guestrin}{Ribeiro
  et~al\mbox{.}}{2018}]%
        {ribeiro2018anchors}
\bibfield{author}{\bibinfo{person}{Marco~Tulio Ribeiro},
  \bibinfo{person}{Sameer Singh}, {and} \bibinfo{person}{Carlos Guestrin}.}
  \bibinfo{year}{2018}\natexlab{}.
\newblock \showarticletitle{Anchors: High-Precision Model-Agnostic
  Explanations.}. In \bibinfo{booktitle}{\emph{Proceedings of the 32nd AAAI
  International Conference on Artificial Intelligence}},
  Vol.~\bibinfo{volume}{18}. \bibinfo{pages}{1527--1535}.
\newblock


\bibitem[\protect\citeauthoryear{R{\"u}ping}{R{\"u}ping}{2006}]%
        {ruping2006learning}
\bibfield{author}{\bibinfo{person}{Stefan R{\"u}ping}.}
  \bibinfo{year}{2006}\natexlab{}.
\newblock \showarticletitle{Learning interpretable models}.
\newblock  (\bibinfo{year}{2006}).
\newblock
\urldef\tempurl%
\url{http://dx.doi.org/10.17877/DE290R-8863}
\showURL{%
\tempurl}


\bibitem[\protect\citeauthoryear{Schmidt and Biessmann}{Schmidt and
  Biessmann}{2019}]%
        {schmidt2019quantifying}
\bibfield{author}{\bibinfo{person}{Philipp Schmidt} {and}
  \bibinfo{person}{Felix Biessmann}.} \bibinfo{year}{2019}\natexlab{}.
\newblock \showarticletitle{Quantifying interpretability and trust in machine
  learning systems}.
\newblock \bibinfo{journal}{\emph{AAAI-19 Workshop on Network Interpretability
  for Deep Learning}} (\bibinfo{year}{2019}).
\newblock


\bibitem[\protect\citeauthoryear{Shin}{Shin}{2021}]%
        {shin2021effects}
\bibfield{author}{\bibinfo{person}{Donghee Shin}.}
  \bibinfo{year}{2021}\natexlab{}.
\newblock \showarticletitle{The effects of explainability and causability on
  perception, trust, and acceptance: Implications for explainable AI}.
\newblock \bibinfo{journal}{\emph{International Journal of Human-Computer
  Studies}}  \bibinfo{volume}{146} (\bibinfo{year}{2021}),
  \bibinfo{pages}{102551}.
\newblock


\bibitem[\protect\citeauthoryear{Shrikumar, Greenside, and Kundaje}{Shrikumar
  et~al\mbox{.}}{2017}]%
        {shrikumar2017learning}
\bibfield{author}{\bibinfo{person}{Avanti Shrikumar}, \bibinfo{person}{Peyton
  Greenside}, {and} \bibinfo{person}{Anshul Kundaje}.}
  \bibinfo{year}{2017}\natexlab{}.
\newblock \showarticletitle{Learning Important Features through Propagating
  Activation Differences}. In \bibinfo{booktitle}{\emph{Proceedings of the 34th
  International Conference on Machine Learning - Volume 70}} (Sydney, NSW,
  Australia) \emph{(\bibinfo{series}{ICML'17})}. \bibinfo{publisher}{JMLR.org},
  \bibinfo{address}{Sydney, NSW, Australia}, \bibinfo{pages}{3145–3153}.
\newblock


\bibitem[\protect\citeauthoryear{Sokol and Flach}{Sokol and Flach}{2020}]%
        {sokol2020explainability}
\bibfield{author}{\bibinfo{person}{Kacper Sokol} {and} \bibinfo{person}{Peter
  Flach}.} \bibinfo{year}{2020}\natexlab{}.
\newblock \showarticletitle{Explainability Fact Sheets: A Framework for
  Systematic Assessment of Explainable Approaches}. In
  \bibinfo{booktitle}{\emph{Proceedings of the 2020 Conference on Fairness,
  Accountability, and Transparency}} (Barcelona, Spain)
  \emph{(\bibinfo{series}{FAT* '20})}. \bibinfo{publisher}{Association for
  Computing Machinery}, \bibinfo{address}{New York, NY, USA},
  \bibinfo{pages}{56–67}.
\newblock
\showISBNx{9781450369367}
\urldef\tempurl%
\url{https://doi.org/10.1145/3351095.3372870}
\showDOI{\tempurl}


\bibitem[\protect\citeauthoryear{Spinner, Schlegel, Sch{\"a}fer, and
  El-Assady}{Spinner et~al\mbox{.}}{2019}]%
        {spinner2019explainer}
\bibfield{author}{\bibinfo{person}{Thilo Spinner}, \bibinfo{person}{Udo
  Schlegel}, \bibinfo{person}{Hanna Sch{\"a}fer}, {and}
  \bibinfo{person}{Mennatallah El-Assady}.} \bibinfo{year}{2019}\natexlab{}.
\newblock \showarticletitle{explAIner: A visual analytics framework for
  interactive and explainable machine learning}.
\newblock \bibinfo{journal}{\emph{IEEE transactions on visualization and
  computer graphics}} \bibinfo{volume}{26}, \bibinfo{number}{1}
  (\bibinfo{year}{2019}), \bibinfo{pages}{1064--1074}.
\newblock


\bibitem[\protect\citeauthoryear{Tenney, Wexler, Bastings, Bolukbasi, Coenen,
  Gehrmann, Jiang, Pushkarna, Radebaugh, Reif, and Yuan}{Tenney
  et~al\mbox{.}}{2020}]%
        {tenney2020language}
\bibfield{author}{\bibinfo{person}{Ian Tenney}, \bibinfo{person}{James Wexler},
  \bibinfo{person}{Jasmijn Bastings}, \bibinfo{person}{Tolga Bolukbasi},
  \bibinfo{person}{Andy Coenen}, \bibinfo{person}{Sebastian Gehrmann},
  \bibinfo{person}{Ellen Jiang}, \bibinfo{person}{Mahima Pushkarna},
  \bibinfo{person}{Carey Radebaugh}, \bibinfo{person}{Emily Reif}, {and}
  \bibinfo{person}{Ann Yuan}.} \bibinfo{year}{2020}\natexlab{}.
\newblock \showarticletitle{The Language Interpretability Tool: Extensible,
  Interactive Visualizations and Analysis for {NLP} Models}.
\newblock  (\bibinfo{year}{2020}), \bibinfo{pages}{107--118}.
\newblock
\urldef\tempurl%
\url{https://www.aclweb.org/anthology/2020.emnlp-demos.15}
\showURL{%
\tempurl}


\bibitem[\protect\citeauthoryear{Tonekaboni, Joshi, McCradden, and
  Goldenberg}{Tonekaboni et~al\mbox{.}}{2019}]%
        {tonekaboni2019clinicians}
\bibfield{author}{\bibinfo{person}{Sana Tonekaboni}, \bibinfo{person}{Shalmali
  Joshi}, \bibinfo{person}{Melissa~D. McCradden}, {and} \bibinfo{person}{Anna
  Goldenberg}.} \bibinfo{year}{2019}\natexlab{}.
\newblock \showarticletitle{What Clinicians Want: Contextualizing Explainable
  Machine Learning for Clinical End Use}. In
  \bibinfo{booktitle}{\emph{Proceedings of the 4th Machine Learning for
  Healthcare Conference}} \emph{(\bibinfo{series}{Proceedings of Machine
  Learning Research}, Vol.~\bibinfo{volume}{106})},
  \bibfield{editor}{\bibinfo{person}{Finale Doshi-Velez}, \bibinfo{person}{Jim
  Fackler}, \bibinfo{person}{Ken Jung}, \bibinfo{person}{David Kale},
  \bibinfo{person}{Rajesh Ranganath}, \bibinfo{person}{Byron Wallace}, {and}
  \bibinfo{person}{Jenna Wiens}} (Eds.). \bibinfo{publisher}{PMLR},
  \bibinfo{address}{Ann Arbor, Michigan}, \bibinfo{pages}{359--380}.
\newblock
\urldef\tempurl%
\url{http://proceedings.mlr.press/v106/tonekaboni19a.html}
\showURL{%
\tempurl}


\bibitem[\protect\citeauthoryear{Wang, Yang, Abdul, and Lim}{Wang
  et~al\mbox{.}}{2019}]%
        {wang2019designing}
\bibfield{author}{\bibinfo{person}{Danding Wang}, \bibinfo{person}{Qian Yang},
  \bibinfo{person}{Ashraf Abdul}, {and} \bibinfo{person}{Brian~Y. Lim}.}
  \bibinfo{year}{2019}\natexlab{}.
\newblock \bibinfo{booktitle}{\emph{Designing Theory-Driven User-Centric
  Explainable AI}}.
\newblock \bibinfo{publisher}{Association for Computing Machinery},
  \bibinfo{address}{New York, NY, USA}, \bibinfo{pages}{1–15}.
\newblock
\showISBNx{9781450359702}
\urldef\tempurl%
\url{https://doi.org/10.1145/3290605.3300831}
\showURL{%
\tempurl}


\bibitem[\protect\citeauthoryear{Zhou, Gandomi, Chen, and Holzinger}{Zhou
  et~al\mbox{.}}{2021}]%
        {zhou2021evaluating}
\bibfield{author}{\bibinfo{person}{Jianlong Zhou}, \bibinfo{person}{Amir~H
  Gandomi}, \bibinfo{person}{Fang Chen}, {and} \bibinfo{person}{Andreas
  Holzinger}.} \bibinfo{year}{2021}\natexlab{}.
\newblock \showarticletitle{Evaluating the quality of machine learning
  explanations: A survey on methods and metrics}.
\newblock \bibinfo{journal}{\emph{Electronics}} \bibinfo{volume}{10},
  \bibinfo{number}{5} (\bibinfo{year}{2021}), \bibinfo{pages}{593}.
\newblock


\end{thebibliography}

%\appendix
%\input{sections/appendix}

\end{document}